\documentclass[11pt,a4paper]{article}

\usepackage{amsmath, amsfonts, amssymb, amsthm, xspace, color}
\usepackage{booktabs} % For formal tables

\usepackage{subfigure, multirow, tabularx} % For subfigure
\usepackage{booktabs} % For formal tables
\usepackage{enumitem}
\usepackage{flushend}
\usepackage[T1]{fontenc}
\usepackage{epstopdf}
\usepackage{balance}
\usepackage{graphicx}
\usepackage[noend,ruled,linesnumbered]{algorithm2e} % For algorithm2e
\usepackage{microtype}
\usepackage{url}
\usepackage{wrapfig,lipsum}
\usepackage{makecell}
\usepackage{wrapfig}

\newcommand{\hide}[1]{} %hide

\newcommand{\etc}{\emph{etc.}\xspace} % and others
\newcommand{\ie}{\emph{i.e.}\xspace} % that is
\newcommand{\eg}{e.g.\xspace} % for example
\newcommand{\nop}[1]{}
\newcommand{\mquote}[1]{{``\emph{#1}''}}

\newtheorem{thm:def}{Definition}
\newtheorem{thm:eg}{Example}
\newtheorem{thm:lem}{Lemma}
\newtheorem{thm:obs}{Observation}
\newtheorem{thm:req}{Requirement}
\newtheorem{thm:prop}{Proposition}
\newtheorem{thm:principle}{Principle}
\newtheorem{thm:hypo}{Hypothesis}
\newtheorem{thm:thm}{Theorem}
\newtheorem{thm:corollary}{Corollary}

\newcommand{\pair}[1]{\langle #1 \rangle}			% use instead of $\langle x \rangle$

\def \P {\mathbf{P}}

\def \D {\mathcal{D}}

\def \G {\mathcal{G}}

\def \K {\mathcal{K}}
\def \M {\mathcal{M}}

\def \S {\mathcal{S}}

\def \V {\mathcal{V}}

\newcommand{\SynSetExpan}{\mbox{SynSetExpan}\xspace}
\newcommand{\SynSetExpanBf}{\mbox{\textbf{SynSetExpan}}\xspace}
\newcommand{\SynSetExpanNoSYN}{\mbox{SynSetExpan-NoSYN}\xspace}

\newcommand{\SynSetExpanNoFT}{\mbox{SynSetExpan-NoFT}\xspace}
\newcommand{\SynSetExpanNoFTBf}{\mbox{\textbf{SynSetExpan-NoFT}}\xspace}

\newcommand{\Dataset}{\mbox{SE2}\xspace}
\newcommand{\DatasetBf}{\mbox{\textbf{SE2}}\xspace}

\definecolor{midnightgreen}{rgb}{0.0, 0.29, 0.33}
\definecolor{orange}{RGB}{255,127,0}

\usepackage[hyperref]{emnlp2020}
\usepackage{times}
\usepackage{latexsym}

\usepackage{microtype}

\aclfinalcopy 
\def\aclpaperid{1070}

\title{SynSetExpan: An Iterative Framework for Joint Entity \\Set Expansion and Synonym Discovery}

\author{Jiaming Shen$^{1\star}$, Wenda Qiu$^{1\star}$, Jingbo Shang$^{2}$, Michelle Vanni$^{3}$, Xiang Ren$^{4}$, Jiawei Han$^1$ \\
\small  $^1$University of Illinois Urbana-Champaign, IL, USA, $^2$University of California San Diego, CA, USA \\
\small $^3$U.S. Army Research Laboratory, MD, USA, $^4$University of Southern California, CA, USA \\
\footnotesize $^1$\{js2, qiuwenda, hanj\}@illinois.edu $\quad$ $^2$jshang@ucsd.edu $\quad$ \footnotesize$^4$michelle.t.vanni.civ@mail.mil $\quad$ $^4$xiangren@usc.edu  \\
}

\begin{document}
\maketitle

{
\renewcommand{\thefootnote}{\fnsymbol{footnote}}
\footnotetext[1]{Equal Contributions.}
}

\begin{abstract}
	%!TEX root = main.tex
% UTF-8 encoding

Entity set expansion and synonym discovery are two critical NLP tasks. 
Previous studies accomplish them separately, without exploring their interdependences.
In this work, we hypothesize that these two tasks are tightly coupled because \emph{two synonymous entities tend to have similar likelihoods of belonging to various semantic classes}.
This motivates us to design \SynSetExpan, a novel framework that enables two tasks to mutually enhance each other.
\SynSetExpan uses a synonym discovery model to include popular entities' infrequent synonyms into the set, which boosts the set expansion recall. 
Meanwhile, the set expansion model, being able to determine whether an entity belongs to a semantic class, can generate pseudo training data to fine-tune the synonym discovery model towards better accuracy.
To facilitate the research on studying the interplays of these two tasks, we create the first large-scale \underline{S}ynonym-\underline{E}nhanced \underline{S}et \underline{E}xpansion (\Dataset) dataset via crowdsourcing.
Extensive experiments on the \Dataset dataset and previous benchmarks demonstrate the effectiveness of \SynSetExpan for both entity set expansion and synonym discovery tasks.
\end{abstract}

%!TEX root = main.tex
% UTF-8 encoding
\section{Introduction}\label{sec:intro}

Entity set expansion (ESE) aims to expand a small set of seed entities (\eg, \{\mquote{United States}, \mquote{Canada}\}) into a larger set of entities that belong to the same semantic class (\ie, \texttt{Country}).
Entity synonym discovery (ESD) intends to group all terms in a vocabulary that refer to the same real-world entity (\eg, \mquote{America} and \mquote{USA} refer to the same country) into a synonym set (hence called a \emph{synset}).
Those discovered entities and synsets include rich knowledge and can benefit many downstream applications such as semantic search~\cite{Xiong2017ExplicitSR}, taxonomy construction~\cite{Shen2018HiExpanTT}, and online education~\cite{Yu2019CourseCE}. 

% Task Formulation
\begin{figure}[!t]
  \centering
  \centerline{\includegraphics[width=0.45\textwidth]{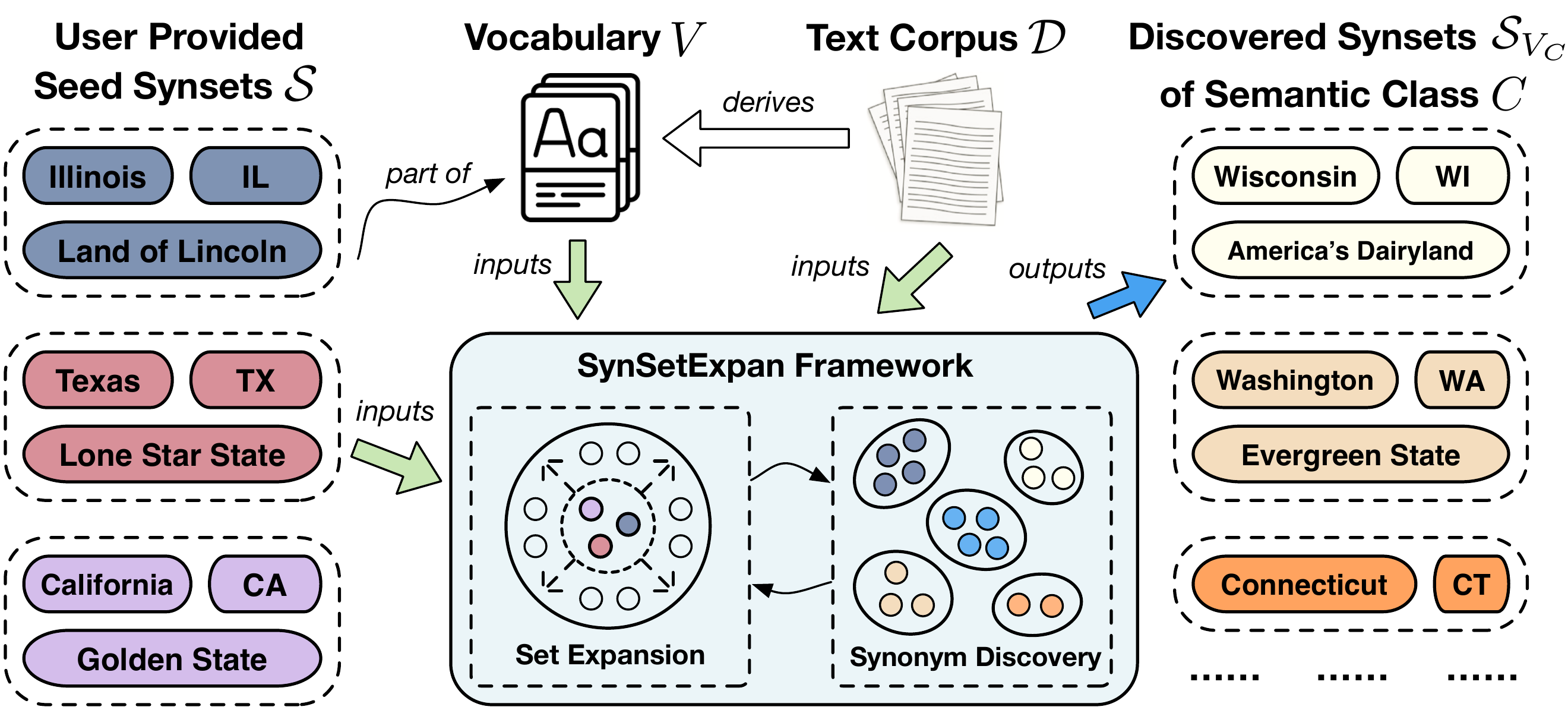}}
  \vspace{-0.2cm}
  \caption{An illustrative example of joint entity set expansion and synonym discovery.}
  \label{fig:task}
  \vspace{-0.3cm}
\end{figure}

Previous studies regard ESE and ESD as two independent tasks. 
Many ESE methods~\cite{Mamou2018TermSE,Yan2019LearningTB,Huang2020GuidingCS,Zhang2020ESELM,Zhu2020FUSEMS} are developed to iteratively select and add the most confident entities into the set.
A core challenge for ESE is to find those infrequent long-tail entities in the target semantic class (\eg, \mquote{Lone Star State} in the class \texttt{US\_States}) while filtering out false positive entities from other related classes (\eg, \mquote{Austin} and \mquote{Dallas} in the class \texttt{City}) as they will cause semantic shift to the set.
Meanwhile, various ESD methods~\cite{Qu2017AutomaticSD,Ustalov2017FightingWT,Wang2019SurfConSD,Shen2019SynSetMine} combine string-level features with embedding features to find a query term's synonyms from a given vocabulary or to cluster all vocabulary terms into synsets.
A major challenge here is to combine those features with limited supervisions in a way that works for entities from all semantic classes.
Another challenge is how to scale a ESD method to a large, extensive vocabulary that contains terms of varied qualities. 

To address the above challenges, we hypothesize that ESE and ESD are two tightly coupled tasks and can mutually enhance each other because \emph{two synonymous entities tend to have similar likelihoods of belonging to various semantic classes and vice versa}.
This hypothesis implies that (1) knowing the class membership of one entity enables us to infer the class membership of all its synonyms, and (2) two entities can be synonyms only if they belong to the same semantic class.
For example, we may expand the \texttt{US\_States} class from a seed set \{\mquote{Illinois}, \mquote{Texas}, \mquote{California}\}. 
An ESE model can find frequent state full names (\eg, \mquote{Wisconsin}, \mquote{Connecticut}) but may miss those infrequent entities (\eg, \mquote{Lone Star State} and \mquote{Golden State}).
However, an ESD model may predict \mquote{Lone Star State} is the synonym of \mquote{Texas} and \mquote{Golden State} is synonymous to \mquote{California} and directly adds them into the expanded set, which shows synonym information help set expansion.
Meanwhile, from the ESE model outputs, we may infer $\langle$\mquote{Wisconsin}, \mquote{WI}$\rangle$ is a synonymous pair while $\langle$\mquote{Connecticut}, \mquote{SC}$\rangle$ is not, and use them to fine-tune an ESD model on the fly.
This relieves the burden of using one single ESD model for all semantic classes and improves the ESD model's inference efficiency because we refine the synonym search space from the entire vocabulary to only the ESE model outputs.

In this study, we propose \SynSetExpan, a novel framework jointly conducting two tasks (c.f. Fig.~\ref{fig:task}).
To better leverage the limited supervision signals in seeds, we design \SynSetExpan as an iterative framework consisting of two components: 
(1) \emph{a ESE model} that ranks entities based on their probabilities of belonging to the target semantic class, and 
(2) \emph{a ESD model} that returns the probability of two entities being synonyms.
In each iteration, we first apply the ESE model to obtain an entity rank list from which we derive a set of \emph{pseudo training} data to fine-tune the ESD model. 
Then, we use this fine-tuned model to find synonyms of entities in the currently expanded set and adjust the above rank list. 
Finally, we add top-ranked entities in the adjusted rank list into the currently expanded set and start the next iteration. 
After the iterative process ends, we construct a synonym graph from the last iteration's output and extract entity synsets (including singleton synsets) as graph communities.

As previous ESE datasets are too small and contain no synonym information for evaluating our hypothesis, we create the first \underline{S}ynonym \underline{E}nhanced \underline{S}et \underline{E}xpansion (\Dataset) benchmark dataset via crowdsourcing.
This new dataset\footnote{\small \url{http://bit.ly/SE2-dataset}.} is one magnitude larger than previous benchmarks.
It contains a corpus of the entire Wikipedia, a vocabulary of 1.5 million terms, and 1200 seed queries from 60 semantic classes of 6 different types (\eg, Person, Location, Organization, \etc). 

\smallskip
\noindent \textbf{Contributions.}~
In summary, this study makes the following contributions:
(1) we hypothesize that ESE and ESD can mutually enhance each other and propose a novel framework \SynSetExpan to jointly conduct two tasks;
(2) we construct a new large-scale dataset \Dataset that supports fair comparison across different methods and facilitates future research on both tasks; and
(3) we conduct extensive experiments to verify our hypothesis and show the effectiveness of \SynSetExpan on both tasks. 

%!TEX root = main.tex
% UTF-8 encoding
\section{Problem Formulation}\label{sec:problem}

%% Figure: Framework Overview
\begin{figure*}[!t]
  \centering
  \centerline{\includegraphics[width=0.98\textwidth]{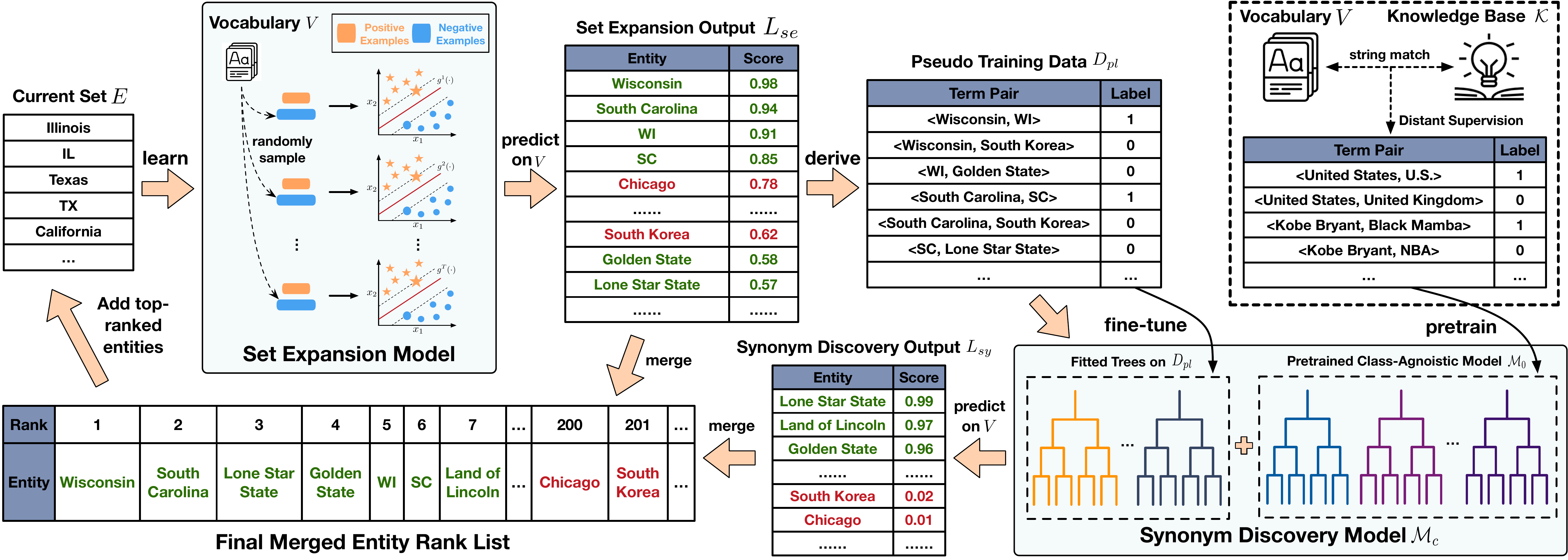}}
  \vspace{-0.2cm}
  \caption{Overview of one iteration in our proposed \SynSetExpan framework. Starting from the current set $E$, we first run a set expansion model to obtain an entity rank list $L_{se}$ based on which we generate pseudo training data $D_{pl}$ to fine-tune a generic synonym discovery model $\mathcal{M}_0$. We then apply this fine-tuned model to get a new rank list $L_{sy}$; merge it with $L_{se}$ to obtain the final entity rank list, and add top ranked entities into the current set $E$.}
  \label{fig:framework}
  \vspace{-0.3cm}
\end{figure*}

We first introduce important concepts in this work, and then present our problem formulation. 
A \textbf{term} is a string (\ie, a word or a phrase) that refers to a real-world entity\footnote{\small In this work, we use ``term'' and ``entity'' interchangeably.}. 
An \textbf{entity synset} is a set of terms that can be used to refer to the same real-world entity. 
For example, both ``USA'' and ``America'' can refer to the entity \emph{United States} and thus compose an entity synset.
We allow the singleton synset and a term may locate in multiple synsets due to its ambiguity.
A \textbf{semantic class} is a set of entities that share a common characteristic and a \textbf{vocabulary} is a term list that can be either provided by users or derived from a corpus.

\smallskip
\noindent \textbf{Problem Formulation.} Given (1) a text corpus $\D$, (2) a vocabulary $\V$ derived from $\D$, and (3) a seed set of user-provided entity synonym sets $\S_0$ that belong to the same semantic class $C$, we aim to (1) select a subset of entities $\V_{C}$ from $\V$ that all belong to $C$; and (2) clusters all terms in $\V_{C}$ into entity synsets $\S_{\V_{C}}$ where the union of all clusters is equal to $\V_{C}$.
In other words, we expand the seed set $\S_0$ into a more complete set of entity synsets $\S_0 \cup \S_{\V_{C}}$ that belong to the same semantic class $C$. 
A concrete example is presented in Fig.~\ref{fig:task}.

%!TEX root = main.tex
% UTF-8 encoding
\section{The SynSetExpan Framework}\label{sec:method}

In this study, we hypothesize that entity set expansion and synonym discovery are two tightly coupled tasks and can mutually enhance each other.
\begin{thm:hypo}\label{thm:hypo1}
Two synonymous entities tend to have similar likelihoods of belonging to various semantic classes and vice versa.
\end{thm:hypo}
The above hypothesis has two implications. 
First, if two entities $e_{i}$ and $e_{j}$ are synonyms and $e_{i}$ belongs to semantic class $C$, $e_{j}$ likely also belongs to class $C$ even if it is currently outside $C$.
This reveals how synonym information can help set expansion by directly introducing popular entities' infrequent synonyms into the set and thus increasing the expansion recall.
The second implication is that if two entities are \emph{not} from the same class $C$, then they are likely not synonyms. 
This shows how set expansion can help synonym discovery by restricting the synonym search space to set expansion outputs and generating additional training data to fine tune the synonym discovery model. 

At the beginning, when we only have limited seed information, this hypothesis may not be directly applicable as we do not have complete knowledge of either entity class memberships or entity synonyms.
Therefore, we design our \SynSetExpan as an iterative framework, shown in Fig.~\ref{fig:framework}.

\smallskip
\noindent \textbf{Framework Overview.}
Before the iterative process starts, we first learn a general synonym discovery model $\M_0$ using distant supervision from a knowledge base (c.f. Sect.~\ref{subsec:synset_model}). 
Then, in each iteration, we learn a set expansion model based on the currently expanded set $E$ (initialized as all entities in user-provided seed synsets $\S_0$) and apply it to obtain a rank list of entities in $\V$, denoted as $L_{se}$ (c.f. Sect.~\ref{subsec:setexpan_model}).
Next, we generate \emph{pseudo training data} from $L_{se}$ and use it to construct a new \emph{class-dependent} synonym discovery model $\M_c$ by fine-tuning $\M_0$.
After that, for each entity in $\V$, we apply $\M_c$ to predict its probability of being the synonym of \emph{at least one} entity in $E$ and use such synonym information to adjust $L_{se}$ (c.f. Sect.~\ref{subsec:mutual_enhance}).
Finally, we add top-ranked entities in the adjusted rank list into the current set and start the next iteration. 
After the iterative process ends, we identify entity synsets from the final iteration's output using a graph-based clustering method (c.f. Sect.~\ref{subsec:synset_construct}).

%%% Synonym Discovery Model
\subsection{Proposed Synonym Discovery Model}\label{subsec:synset_model}

Given a pair of entities, our synonym discovery model returns the probability that they are synonymous. 
We use two types of features for entity pairs\footnote{\small We list all features in supplementary materials Section A.}:
(1) \emph{lexical features} based on entity surface names (\eg, Jaro-Winkler similarity~\cite{Wang2019SurfConSD}, token edit distance~\cite{Fei2019HierarchicalMW}, etc), and
(2) \emph{semantic features} based on entity embeddings (\eg, cosine similarity between two entities' SkipGram embeddings).
As these feature values have different scales, we use a tree-based boosting model XGBoost~\cite{Chen2016XGBoostAS} to predict whether two entities are synonyms.
Another advantage of XGBoost is that it is an additive model and supports incremental model fine-tuning. 
We will discuss how to use set expansion results to fine-tune a synonym discovery model in Sect.~\ref{subsec:setexpan_model}. 

To learn the synonym discovery model, we first acquire distant supervision data by matching each term in the vocabulary $\V$ with the canonical name of one entity (with its unique ID) in a knowledge base (KB).
If two terms are matched to the same entity in KB and their embedding similarity is larger than 0.5, we treat them as synonyms.
To generate a non-synonymous term pair, we follow the same ``mixture'' sampling strategy proposed in~\cite{Shen2019SynSetMine}, that is, 50\% of negative pairs come from random sampling and the other 50\% of negative pairs are those ``hard'' negatives which are required to share at least one token. 
Some concrete examples are shown in Fig.~\ref{fig:framework}. 
Finally, based on such generated distant supervision data, we train our XGBoost-based synonym discovery model using binary cross entropy loss.

%%% Set Expansion Model
\subsection{Proposed Set Expansion Model}\label{subsec:setexpan_model}

Given a set of seed entities $E_0$ from a semantic class $C$, we aim to learn a set expansion model that can predict the probability of a new entity (term) $e_i \in \V$ belonging to the same class $C$, \ie, $\mathbf{P}(e_i \in C)$. 
We follow previous studies~\cite{Melamud2016TheRO,Mamou2018TermSE2} to represent each entity using a set of 6 embeddings learned on the given corpus $\mathcal{D}$, including SkipGram, CBOW in word2vec~\cite{Mikolov2013DistributedRO}, fastText~\cite{bojanowski2016enriching}, SetExpander~\cite{Mamou2018TermSE}, JoSE~\cite{Meng2019SEmed} and averaged BERT contextualized embeddings~\cite{Devlin2019BERTPO}. 
Given the bag-of-embedding representation $[\mathbf{f}^{1}(e_i), \mathbf{f}^{2}(e_i), \dots, \mathbf{f}^{B}(e_i)]$ of entity $e_i$ and the seed set $E_0$, we define the entity feature $\mathbf{x}_{i} = \|_{b=1}^{6} \|_{j=1}^{|E_0|} \left[\sqrt{d_{ij}^b}, ~d_{ij}^b, ~(d_{ij}^{b})^{2}\right]$,
where ``$\|$'' represents the concatenation operation, and $d_{ij}^b = \text{cos}(\mathbf{f}^{b}(e_i), \mathbf{f}^{b}(e_j))$ is the cosine similarity between two embedding vectors. 
One challenge of learning the set expansion model is the lack of supervision signals --- we only have a few ``positive'' examples (\ie, entities belonging to the target class) and no ``negative''  examples. 
To solve this challenge, we observe that \emph{the size of target class is usually much smaller than the vocabulary size}. 
This means if we randomly select one entity from the vocabulary, most likely it will not belong to the target semantic class. 
Therefore, we can construct a set of $|E_0| \times K$ negative examples by random sampling.
We also test selecting only entities that have a low embedding similarity with the entities in the current set. 
However, our experiment shows this restricted sampling does not improve the performance.
Therefore, we choose to use the simple yet effective ``random sampling'' approach and refer to $K$ as ``negative sampling ratio''. 
Given a total of $|E_0| \times (K+1)$ examples, we learn a SVM classifier $g(\cdot)$ based on the above defined entity features. 

To further improve set expansion quality, we repeat the above process $T$ times (\ie, randomly sample $T$ different sets of $|E_0| \times K$ negative examples for learning $T$ separate classifiers $\{g^{t}(\cdot)\}|_{t=1}^{T}$) and construct an ensemble classifier.
The final classifier predicts the probability of an entity $e_i$ belonging to the class $C$ by averaging all individual classifiers' outputs (\ie, $\mathbf{P}(e_i \in C) = \frac{1}{T}\sum_{t=1}^{T} g^{t}(e_i)$.
Finally, we rank all entities in the vocabulary based on their predicted probabilities.

\subsection{Two Models' Mutual Enhancements}\label{subsec:mutual_enhance}

\noindent \textbf{Set Expansion Enhanced Synonym Discovery.}
In each iteration, we generate a set of \emph{pseudo training} data $D_{pl}$ from the ESE model output $L_{se}$, to fine-tune the general synonym discovery model $\M_0$. 
Specifically, we add an entity pair $\pair{e_x, e_y}$ into $D_{pl}$ with label 1, if they are among the top 100 entities in $L_{se}$ and $\M_{0}(e_x, e_y) \geq 0.9$.
For each positive pair $\pair{e_x, e_y}$, we generate $N$ negative pairs by randomly selecting $\lceil N/2 \rceil$ entities from $L_{se}$ whose set expansion output probabilities are less than 0.5 and pairing them with both $e_x$ and $e_y$.
The intuition is that those randomly selected entities likely come from different semantic classes with entity $e_x$ and $e_y$, and thus based on our hypothesis, they are unlikely to be synonyms. 
After obtaining $D_{pl}$, we fine-tune model $\M_0$ by fitting $H$ additional trees on $D_{pl}$ and incorporate them into the existing bag of trees in $\M_0$. 
We discuss the detailed choices of $N$ and $H$ in the experiment. 

% Algorithm: Overall Framework
 \small
 \begin{algorithm}[!t]
   \caption{\SynSetExpan Framework.}
   \label{algo:synsetexpan_framework}
   \KwIn{\small
   A seed set $\S_0$, a vocabulary $\V$, a knowledge base $\K$, maximum iteration number \emph{max$\underline{\hspace{0.05in}}$iter}, maximum size of expanded set $Z$, and model hyper-parameters $\{K, T, N, H\}$.
   }
   \KwOut{\small A complete set of entity synsets $\S_{\V_{C}}$.}
   \small Learn a general ESD model $\M_{0}$ using distant supervision in $\K$\;
   $E \gets $ Union of all synsets in $\S_0$\;
   \For{iter from 1 to max$\underline{\hspace{0.05in}}$iter} {
   	$L_{se} \gets \text{ESEModel}(E, \V, K, T)$\;
	Generate pseudo training data $D_{pl}$ from $L_{se}$\;
	Construct a class-specific ESD model $\M_{c}$ by fine-tuning $\M_{0}$ on $D_{pl}$\;
	Apply $\M_{c}$ on entities in $\V$ and adjust $L_{se}$\;
	Add top $\lceil \frac{Z}{\text{max}\underline{\hspace{0.05in}}\text{iter}} \rceil $ entities in the adjusted rank list into $E$\;
   }
   Construct a synonym graph $\G$ based on final set $E$\;
   $\S_{\V_{C}} \gets \text{Louvain}(\G)$\;
   Return $\S_{\V_{C}}$.
 \end{algorithm}
 \normalsize
 
\vspace{0.1em}
\noindent \textbf{Synonym Enhanced Set Expansion.}
Given a fine-tuned class-specific synonym discovery model $\M_{c}$, the current set $E$, we calculate a new score for each entity $e_{i} \in \V$ as follows:
\begin{equation}
\small
\text{sy-score}(e_i) = \text{max} \{ \M_{c}(e_i, e_j) | e_j \in E \}. 
\end{equation}
The above score measures the probability that $e_{i}$ is the synonym of one entity in $E$.
Based on Hypothesis~\ref{thm:hypo1}, we know an entity with a large sy-score is likely belonging to the target class. 
Therefore, we use a multiplicative measure to combine this sy-score with set expansion model's original output $\P(e_i \in C)$ as follows:
\begin{equation}\label{eq:final_score}
\small
\text{final-score}(e_i) = \sqrt{\P(e_i \in C) \times \text{sy-score}(e_i)}. 
\end{equation}
An entity will have a large sy-score as long as it is the synonym of \emph{one single} entity in $E$.
Such a property is particularly important for capturing long-tail infrequent entities.
For example, suppose we expand \texttt{US\_States} class from a seed set \{\mquote{Illinois}, \mquote{IL}, \mquote{Texas}, \mquote{TX}\}.
The original set expansion model, biased toward popular entities, assigns a low score 0.57 to \mquote{Lone Star State} and a large score 0.78 to \mquote{Chicago}. 
However, the synonym discovery model predicts that, over 99\% probability, \mquote{Lone Star State} is the synonym of \mquote{Texas} and thus has a sy-score 0.99.
Meanwhile, \mquote{Chicago} has no synonym in the seed set and thus has a low sy-score 0.01.
As a result, the final score of \mquote{Lone Star State} is larger than that of \mquote{Chicago}.
Moreover, we emphasize that Eq.~\ref{eq:final_score} uses synonym scores to enhance, not replace, set expansion scores. 
A correct entity $e^{\star}$ that has no synonym in current set $E$ will indeed be ranked after other correct entities that have synonyms in $E$.
However, this is not problematic because (1) all compared entities are correct, and (2) we will not remove $e^{\star}$ from final results because it still outscores other erroneous entities that have the same low sy-score as $e^{\star}$ but much lower set expansion scores. 

\subsection{Synonym Set Construction}\label{subsec:synset_construct}
After the iterative process ends, we have a synonym discovery model $\M_{c}$ that predicts whether two entities are synonymous and an entity list $E$ that includes entities from the same semantic class. 
To further derive entity synsets, we first construct a weighted synonym graph $\G$ where each node $n_i$ represents one entity $e_i \in E$ and each edge $(n_i, n_j)$ with weight $w_{ij}$ indicates $M_{c}(e_i, e_j) = w_{ij}$.
Then, we apply the non-overlapping community detection algorithm Louvain~\cite{Blondel2008FastUO} to find all clusters in $\G$ and treat them as entity synsets. 
Note here we narrow the original full vocabulary $\V$ to set expansion model's final output $E$ based on our hypothesis. 
We summarize our whole framework in Algorithm~\ref{algo:synsetexpan_framework} and discuss its computational complexity in supplementary materials.

%!TEX root = main.tex
% UTF-8 encoding
\section{The SE2 Dataset}\label{sec:dataset}

To verify our hypothesis and evaluate \SynSetExpan framework, we need a dataset that contains a corpus, a vocabulary with labeled synsets, a set of complete semantic classes, and a list of seed queries.
However, to the best of our knowledge, there is no such a public benchmark\footnote{\small More discussions on existing set expansion datasets are available in supplementary materials Section C.}.
Therefore, we build the first \underline{S}ynonym \underline{E}nhanced \underline{S}et \underline{E}xpansion (\DatasetBf) benchmark dataset in this study\footnote{\small More details and analysis can be found in the Section D and E of supplementary materials.}. 

\subsection{Dataset Construction}

We construct the SE2 dataset in four steps.

%% Data Statistics
\begin{table}[t]
\centering
\scalebox{0.85}{
    \small
    \begin{tabular}{ccccc}
        \toprule
        \textbf{Corpus Size} & \textbf{\# Entities} & \textbf{\# Classes} & \textbf{\# Queries} \\
        \midrule
        1.9B Tokens & 1.5M & 60 & 1200 \\
        \bottomrule
    \end{tabular}
}
\vspace{-0.2cm}
\caption{Our SE2 dataset statistics}\label{table:dateset}
\vspace{-0.3cm}
\end{table}

\vspace{0.1em}
\noindent \textbf{1. Corpus and Vocabulary Selection.} 
We use Wikipedia 20171201 dump as our evaluation corpus as it contains a diverse set of semantic classes and enough context information for methods to discover those sets. 
We extract all noun phrases with frequency above $10$ as our selected vocabulary.

\vspace{0.1em}
\noindent \textbf{2. Semantic Class Selection.} 
We identify 60 major semantic classes based on the  \emph{DBpedia-Entity v2}~\cite{Hasibi2017DBpediaEntityVA} and \emph{WikiTable}~\cite{Bhagavatula2015TabELEL} entities found in our corpus.
These 60 classes cover 6 different entity types (\eg, Person, Location, Organization).
As such generated classes may miss some correct entities, we enlarge each class via crowdsourcing in the following step.

\vspace{0.1em}
\noindent \textbf{3. Query Generation and Class Enrichment.} 
We first generate 20 queries for each semantic class.
Then, we aggregate the top 100 results from all baseline methods (c.f. Sect.~\ref{sec:exp}) and obtain 17,400 $\pair{\text{class}, \text{entity}}$ pairs. 
Next, we employ crowdworkers on Amazon Mechanical Turk to check all those pairs.
Workers are asked to view one semantic class and six candidate entities, and to select all entities that belong to the given class. 
On average, workers spend 40 seconds on each task and are paid \$0.1.
All $\pair{\text{class}, \text{entity}}$ pairs are labeled by three workers independently and the inter-annotator agreement is 0.8204, measured by Fleiss's Kappa ($k$).
Finally, we enrich each semantic class $C_j$ by adding the entity $e_i$ whose corresponding pair $\pair{C_j, e_i}$ is labeled ``True'' by at least two workers.

\vspace{0.1em}
\noindent \textbf{4. Synonym Set Curation.} 
To construct synsets in each class, we first run all baseline methods to generate a candidate pool of possible synonymous term pairs. 
Then, we treat those pairs with both terms mapped to the same entity in WikiData as positive pairs and ask two human annotators to label the remaining 7,625 pairs.
The inter-annotator agreement is 0.8431, measured by Fleiss's Kappa.
Then, we construct a synonym graph where each node is a term and each edge connects two synonymous terms. 
Finally, we extract all connected components in this graph and treat them as synsets.

\subsection{Dataset Analysis}

We analyze some properties of \Dataset dataset from the following three aspects.

% Table: Dataset Analysis.
\begin{table}[!t]
\centering
\scalebox{0.75}{
\begin{tabular}{cccc}
\toprule
\textbf{Class Type} & \textbf{ESE} & \textbf{ESD (Lexical)} & \textbf{ESE (Semantic)} \\
\midrule
Location & 0.3789 & 0.2132 & 0.6599 \\
Person & 0.2322 & 0.2874 & 0.5526 \\
Product & 0.0848 & 0.3922 & 0.4811 \\
Facility &0.0744 & 0.2345 & 0.4466 \\
Organization & 0.1555 & 0.2566 & 0.4935 \\ 
Misc & 0.4282 & 0.2743 & 0.5715 \\  
\bottomrule
\end{tabular}}
\vspace*{-0.1cm}
\caption{\label{tbl:data_analysis}Difficulty of each semantic class for entity set expansion (ESE) and entity synonym discovery (ESD).}
\vspace*{-0.3cm}
\end{table}

\vspace{0.1em}
\noindent \textbf{1. Semantic class size.} 
The 60 semantic classes in our \Dataset dataset consist on average 145 entities (with a minimum of 16 and a maximum of 864) for a total of 8697 entities. 
After we grouping these entities into synonym sets, these 60 classes consist of on average 118 synsets (with a minimum of 14 and a maximum of 800) for totally 7090 synsets. 
The average synset size is 1.258 and the maximum size of one synset is 11. 

\vspace{0.1em}
\noindent \textbf{2. Set expansion difficulty of each class.} 
We define the set expansion difficulty of each semantic class as follows:
\begin{equation}\label{eq:set_expansion_difficulty}
\small
\text{Set-Expansion-Difficulty}(C) = \frac{1}{|C|}\sum_{e \in C} \frac{|C - \text{Top}k(e)|}{|C|},
\end{equation}
where Top$k(e)$ represents the set of $k$ most similar entities to entity $e$ in the vocabulary. 
We set $k$ to be 10,000 in this study. 
Intuitively, this metric calculates the average portion of entities in class $C$ that cannot be easily found by another entity in the same class. 
As shown in Table~\ref{tbl:data_analysis}, the most difficult classes are those Location classes\footnote{\small We exclude MISC type because by its definition classes of this type will be very random.} and the easiest ones are Facility classes. 

\vspace{0.1em}
\noindent \textbf{3. Synonym discovery difficulty of each class.} 
We continue to measure the difficulty of finding synonym pairs in each class.
Specifically, we calculate two metrics: (1) \emph{Lexical difficulty} defined as the average Jaro-Winkler distance between the surface names of two synonyms, and (2) \emph{Semantic difficulty} defined as the average cosine distance between two synonymous entities' embeddings. 
Table~\ref{tbl:data_analysis} lists the results. 
We find Product classes have the largest lexical difficulty and Location classes have the largest semantic difficulty.

% Table: Set Expansion Overall Result
    \begin{table*}
    \centering
    \scalebox{0.68}{
    % \small
    \begin{tabular}{lccccccccc}
    \toprule
    \multirow{2}{*}{\textbf{Methods}} & \multicolumn{3}{c}{\textbf{SE2}}   & \multicolumn{3}{c}{\textbf{Wiki}}            & \multicolumn{3}{c}{\textbf{APR}}                          \\
      & MAP@10         & MAP@20         & MAP@50                         & MAP@10         & MAP@20         & MAP@50         & MAP@10         & MAP@20         & MAP@50         \\ 
                               
    \toprule
    Egoset~\cite{Rong2016egoset}               &     0.583      &    0.533       &  0.433   & 0.904          & 0.877          & 0.745          & 0.758          & 0.710          & 0.570          \\ 
    SetExpan~\cite{Shen2017SetExpanCS}      &   0.473        &    0.418       &     0.341         & 0.944            & 0.921          & 0.720          & 0.789          & 0.763          & 0.639          \\ 
    SetExpander~\cite{Mamou2018TermSE}    &    0.520       &  0.475         &    0.397         & 0.499          & 0.439          & 0.321          & 0.287          & 0.208          & 0.120          \\
    MCTS~\cite{Yan2019LearningTB} &    ---       &    ---       & --- & 0.980 & 0.930 & 0.790 & 0.960 & 0.900 & 0.810  \\
    CaSE~\cite{Yu2019CorpusbasedSE}          &  0.534         &  0.497         &     0.420         & 0.897          & 0.806          & 0.588          & 0.619          & 0.494          & 0.330          \\
    SetCoExpan~\cite{Huang2020GuidingCS}   &    ---       &  ---         & --- & 0.976 & 0.964 &  0.905 & 0.933 & 0.915 &  0.830  \\
    CGExpan~\cite{Zhang2020ESELM}  &   0.601       &   0.543        &  0.438 & \textbf{0.995} & \textbf{0.978} & 0.902 & \textbf{0.992} & \textbf{0.990} & 0.955  \\ 
    \midrule
    SynSetExpan-NoSYN    & 0.612      & 0.567  & 0.484    & 0.991 & \textbf{0.978} & \textbf{0.904} & 0.985 & \textbf{0.990} & \textbf{0.960}  \\
    SynSetExpan                & \textbf{0.628}$^{*}$          &  \textbf{0.584}$^{*}$          & \textbf{0.502}$^{*} $    & --- & --- & --- & --- & --- & --- \\ \bottomrule
    \end{tabular}
    }
    \vspace*{-0.2cm}
    \caption{\label{tbl:setexpan_results}\small Set expansion results on three datasets. MCTS and SetCoExpan do not scale to the SE2 dataset. SynSetExpan-Full is inapplicable for Wiki and APR datasets because they contain no synonym information. The superscript $^{*}$ indicates the improvement is statistically significant compared to SynSetExpan-NoSYN.}
    \vspace*{-0.1cm}
    \end{table*}
%!TEX root = main.tex
% UTF-8 encoding
\section{Experiments}\label{sec:exp}
\vspace{-0.1cm}

\subsection{Entity Set Expansion}
\noindent \textbf{Datasets.}
We evaluate \SynSetExpan on three public datasets.
The first two are benchmark datasets widely used in previous studies~\cite{Shen2017SetExpanCS, Yan2019LearningTB, Zhang2020ESELM}: 
(1) \textbf{Wiki}, which contains 8 semantic classes, 40 seed queries, and a subset of English Wikipedia articles, and
(2) \textbf{APR}, which includes 3 semantic classes, 15 seed queries, and all news articles published by Associated Precess and Reuters in 2015. 
Note that these two datasets do not contain synonym information and are used primarily to evaluate our set expansion model performance.
We decide not to augment these two datasets with additional synonym information (as we did in our SE2 dataset) in order to keep the integrity of two existing benchmarks.
The third one is our proposed \Dataset dataset which has 60 semantic classes, 1200 seed queries, and a corpus of 1.9 billion tokens. 
Clearly, our \Dataset is of one magnitude larger than previous benchmarks and covers a wider range of semantic classes. 

\smallskip
\noindent \textbf{Compared Methods.}
We compare the following corpus-based set expansion methods: 
(1) \textbf{EgoSet}~\cite{Rong2016egoset}: A method initially proposed for multifaceted set expansion using skip-grams and word2vec embeddings. 
Here, we treat all extracted entities forming in one set as our queries have little ambiguity.
(2) \textbf{SetExpan}~\cite{Shen2017SetExpanCS}: A bootstrap method that first computes entity similarities based on selected quality contexts and then expands the entity set using rank ensemble.
(3) \textbf{SetExpander}~\cite{Mamou2018TermSE}: A one-time entity ranking method based on multi-context term similarity defined on multiple embeddings. 
(4) \textbf{MCTS}~\cite{Yan2019LearningTB}: A bootstrap method combining the Monte Carlo Tree Search algorithm with a deep similarity network to estimate delayed feedback for pattern evaluation and entity scoring.
(5) \textbf{CaSE}~\cite{Yu2019EfficientCS}: Another one-time entity ranking method using both term embeddings and lexico-syntactic features. 
(6) \textbf{SetCoExpan}~\cite{Huang2020GuidingCS}: A set expansion framework which generates auxiliary sets that are closely related to the target set and leverages them to guide the expansion process. 
(7) \textbf{CGExpan}~\cite{Zhang2020ESELM}: Current state-of-the-art method that generates the target set name by querying a pre-trained language model and utilizes generated names to expand the set. 
(8) \textbf{SynSetExpan}: Our proposed framework which jointly conducts two tasks and enables synonym information to help set expansion. 
(9) \textbf{SynSetExpan-NoSYN}: A variant of our proposed \SynSetExpan framework without the synonym discovery model. 
All implementation details and hyperparameter choices are discussed in supplementary materials Section F.

\smallskip
\noindent \textbf{Evaluation Metrics.}
We follow previous studies and evaluate our results using Mean Average Precision at different top $K$ positions: $\text{MAP@}\textit{K} = \frac{1}{|Q|} \sum_{q \in Q} \text{AP}_{K} (L_{q}, S_{q})$, 
where $Q$ is the set of all seed queries and for each query $q$, we use $\text{AP}_{K} (L_{q}, S_{q})$ to denote the traditional average precision at position $K$ given a ranked list of entities $L_{q}$ and a ground-truth set $S_{q}$. 
To compare the performance of multiple models, we conduct statistical significance test using the two-tailed paired t-test with 99\% confidence level.

\smallskip
\noindent \textbf{Experimental Results.}
We analyze the set expansion performance from the following aspects. 

\vspace{0.1em}
\noindent \textbf{1. Overall Performance.}
Table~\ref{tbl:setexpan_results} presents the overall set expansion results. 
We can see that \SynSetExpanNoSYN achieves comparable performances with the current state-of-the-art methods on Wiki and APR datasets\footnote{\small We feel both CGExpan and our method have reached the performance limit on Wiki and APR as both datasets are relatively small and contain only a few coarse-grained classes.}, and outperforms previous methods on \Dataset dataset, which demonstrates the effectiveness of our set expansion model alone. 
Besides, by comparing \SynSetExpanNoSYN with \SynSetExpan on \Dataset dataset, we show that adding synonym information indeed helps set expansion. 

\begin{figure*}[!t]
  \centering
  \centerline{\includegraphics[width=0.98\textwidth]{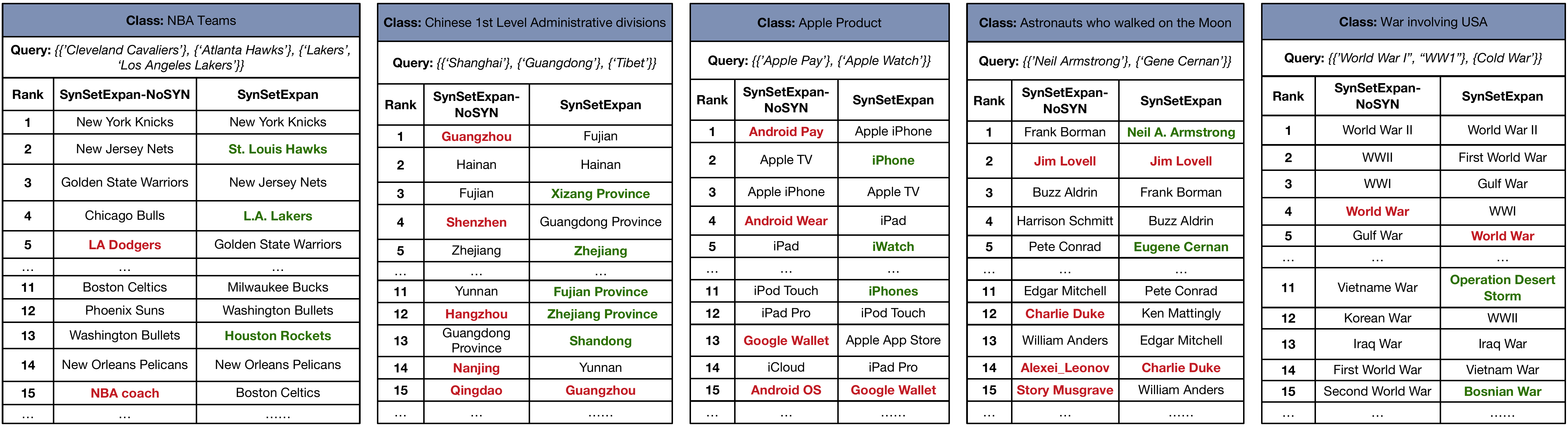}}
  \vspace{-0.3cm}
  \caption{Case studies on entity set expansion. Erroneous entities are colored in red. Entities discovered only by \SynSetExpan in top-20 results are colored in green.}
  \label{fig:set_expan_cases}
  \vspace{-0.1cm}
\end{figure*}

\vspace{0.1em}
\noindent \textbf{2. Fine-grained Performance Analysis.}
To provide a detailed analysis on how \SynSetExpan improves over \SynSetExpanNoSYN, we group semantic classes based on their types and calculate the ratio of classes on which \SynSetExpan outperforms \SynSetExpanNoSYN.
Table~\ref{table:compare_nosyn} shows the results and we can see that on most classes \SynSetExpan is better than \SynSetExpanNoSYN, especially for the MAP@50 metric. 
In Table~\ref{table:compare}, we further analyze the ratio of seed set queries (out of total 1200 queries) on which one method achieves better or the same performance as the other method.
We can see that \SynSetExpan can win on the majority of queries, which further shows that \SynSetExpan can effectively leverage synonym information to enhance set expansion. 

    %% Table: Comparison between SynSetExpan and \SynSetExpanNoSYN
    \begin{table}[!t]
    \centering
    \scalebox{0.95}{
        \small
        \begin{tabular}{cccc}
        \toprule
        \textbf{Class Type} & \textbf{MAP@10} & \textbf{MAP@20}  & \textbf{MAP@50} \\                                   
        \midrule
        Person          & 86.7\%  & 80.0\%  & 93.3\%   \\
        Organization    & 83.3\%  & 83.3\%  & 100\%   \\
        Location        & 69.2\%  & 65.4\%  & 80.8\%   \\
        Facility        & 85.7\%  & 71.4\%  & 100\%   \\
        Product     & 100\%  & 66.7\%  & 100\%   \\
        Misc            & 66.7\%  & 66.7\%  & 100\%   \\
        \midrule
        \textbf{Overall}        & \textbf{78.3\%} & \textbf{71.7\%} & \textbf{90.0\%} \\
        \bottomrule
        \end{tabular}
    }
        \vspace{-0.2cm}
        \caption{\label{table:compare_nosyn}\small Ratio of semantic classes on which \SynSetExpan outperforms \SynSetExpanNoSYN.}
        \vspace{-0.1cm}
    \end{table}
    
    \begin{table}
    \centering
    \scalebox{0.88}{
        \small
        \begin{tabular}{lccc}
            \toprule
            \textbf{\SynSetExpan vs. Other}   & \textbf{MAP@10}     & \textbf{MAP@20}    & \textbf{MAP@50}      \\
            \midrule
            vs. CGExpan         & 78.9\%     & 85.4\%     & 93.8\%        \\ 
            vs. \SynSetExpanNoSYN      & 72.7\%     & 83.0\%      & 91.4\%       \\ 
            \bottomrule
        \end{tabular}
    }
    \vspace{-0.2cm}
    \caption{\label{table:compare} \small Ratio of seed queries from the SE2 dataset on which the first method outperforms the second one.}
    \vspace{-0.1cm}
    \end{table}
    
\vspace{0.1em}
\noindent \textbf{3. Case Studies.}
Figure~\ref{fig:set_expan_cases} shows some expanded semantic classes by \SynSetExpan.
We can see that the set expansion task benefits a lot from the synonym information.
Take the semantic class \texttt{NBA\_Teams} for example, we find \mquote{L.A. Lakers} (\ie, the synonym of \mquote{Los Angeles Lakers}) as well as \mquote{St. Louis Hawks} (\ie, the former name of \mquote{Atlanta Hawks}) and further use them to improve the set expansion result. 
Moreover, by introducing synonyms, we can lower the rank of those erroneous entities (\eg, \mquote{LA Dodgers} and \mquote{NBA coach}).

%% Section 2: Synonym Discovery
\subsection{Synonym Discovery}

\noindent \textbf{Datasets.}
We evaluate \SynSetExpan for synonym discovery task on two datasets: (1) \DatasetBf, which contains 60,186 synonym pairs (3,067 positive pairs and 57,119 negative pairs), and (2) \textbf{PubMed}, a public benchmark used in~\cite{Qu2017AutomaticSD,Shen2019SynSetMine}, which contains 203,648 synonym pairs (10,486 positive pairs and 193,162 negative pairs). 
More details can be found in supplementary materials Section G.1. 
    
% Table: Synonym Discovery Overall Result
    \begin{table}
    \centering
    \scalebox{0.52}{
    % \small
    \begin{tabular}{lcccccc}
    \toprule
    \multirow{2}{*}{\textbf{Method}}   & \multicolumn{3}{c}{\textbf{SE2}}                         & \multicolumn{3}{c}{\textbf{PubMed}}                          \\
                               & AP         & AUC         & F1         & AP         & AUC         & F1         \\ 
                               
    \midrule
        SVM                                     & 0.1870    & 0.8547  & 0.3300  & 0.2250    & 0.8206  & 0.4121 \\
        XGB-S~\cite{Chen2016XGBoostAS}      & 0.7654 & 0.9696  & 0.6389  & 0.5012 & 0.8625  & 0.4968 \\
        XGB-E~\cite{Chen2016XGBoostAS}  & 0.4762 & 0.8750  & 0.4810  & 0.4906 & 0.9190  & 0.5388 \\
        DPE~\cite{Qu2017AutomaticSD}                & 0.7972 & 0.9792  & 0.6392  & 0.6338 & 0.8979  & 0.6038 \\
        SynSetMine~\cite{Shen2019SynSetMine}            & 0.7562 & 0.9782  & 0.6347  & 0.6757 & 0.9453  & 0.6287 \\
    \midrule
        \SynSetExpanNoFT & 0.8197 & 0.9844  & 0.7159  & 0.6615 & 0.9445  & 0.6204 \\
        \SynSetExpan & \textbf{0.8736} & \textbf{0.9953}  & \textbf{0.7592}  & \textbf{0.7152} & \textbf{0.9695}  & \textbf{0.6388} \\
    \bottomrule
    \end{tabular}
    }
    \vspace*{-0.2cm}
    \caption{\label{tbl:synonym_results} Synonym discovery results on both \DatasetBf dataset and PubMed dataset.}
    \vspace*{-0.3cm}
    \end{table}
    
\smallskip
\noindent \textbf{Compared Methods.}
We compare following synonym discovery methods: 
(1) \textbf{SVM}: A classification method trained on given term pair features. We use the same feature set described in Sect.~\ref{subsec:synset_model}.
(2) \textbf{XGBoost}~\cite{Chen2016XGBoostAS}: Another classification method trained on given term pair features. Here, we test its two variants: \textbf{XGB-S} which only leverages lexical features based on entity surface names, and \textbf{XGB-E} which only utilizes entity embedding features. 
(3) \textbf{DPE}~\cite{Qu2017AutomaticSD}: A distantly supervised method integrating embedding features and textual patterns for synonym discovery. 
(4) \textbf{SynSetMine}~\cite{Shen2019SynSetMine}: Another distantly supervised framework that learns to represent the entire entity synonym set. 
(5) \SynSetExpanBf: Our proposed framework that fine-tunes synonym discovery model using set expansion results.
(6) \SynSetExpanNoFTBf: A variant of \SynSetExpan without using the model fine-tuning. 
More implementation details and hyper-parameter choices are discussed in supplementary materials Section G. 

%% Figure: Synonym Case Studies
\begin{figure}[!t]
  \centering
  \centerline{\includegraphics[width=0.48\textwidth]{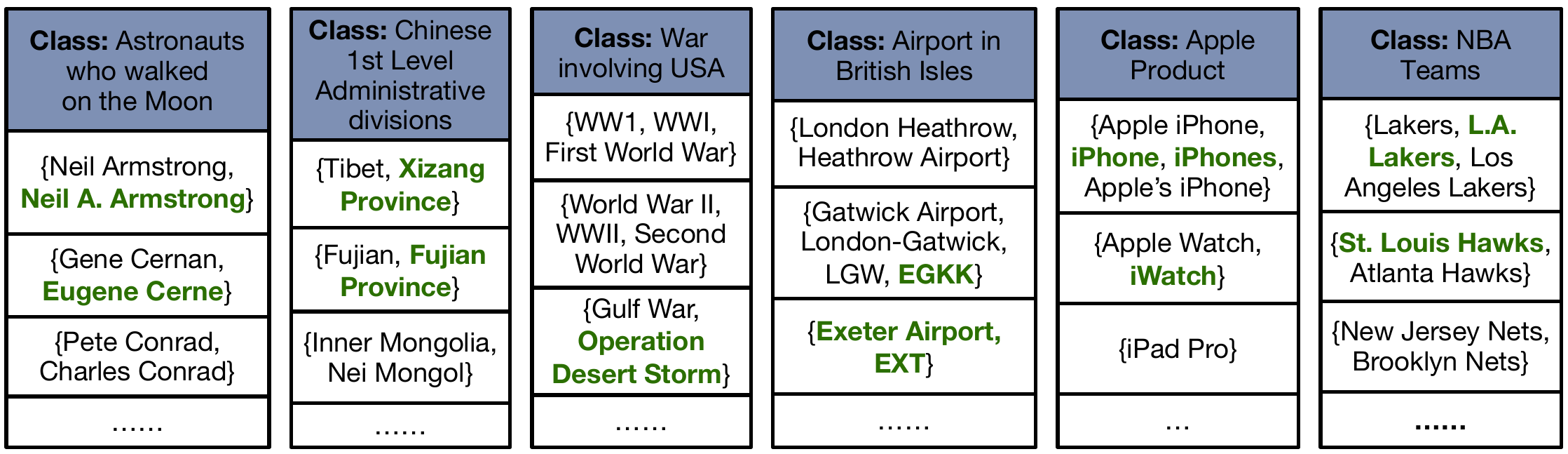}}
  \vspace{-0.1cm}
  \caption{Case studies on synonym discovery. Entities discovered only by \SynSetExpan are colored in green.}
  \label{fig:synonym_cases}
  \vspace{-0.1cm}
\end{figure}

\smallskip
\noindent \textbf{Evaluation Metrics.}
As all compared methods output the probability of two input terms being synonyms, we first use two threshold-free metrics for evaluation --- Average Precision (\textbf{AP}) and Area Under the ROC Curve (\textbf{AUC}). 
Second, we transform the output probability to a binary decision using threshold 0.5 and evaluate the model performance using standard \textbf{F1} score. 
    
\smallskip
\noindent \textbf{Experimental Results.}
Table~\ref{tbl:synonym_results} shows the overall synonym discovery results.
First, we can see that the \SynSetExpanNoFT model can outperform both XGB-S and XGB-E methods significantly, which shows the importance of using both types of features for predicting synonyms. 
Second, we find that \SynSetExpan can further improve \SynSetExpanNoFT via model fine-tuning, which demonstrates that set expansion can help synonym discovery.  
Finally, we notice that our \SynSetExpan framework, with the fine-tuning mechanism enabled, can achieve the best performance across all evaluation metrics.
In Figure~\ref{fig:synonym_cases}, we show some synsets discovered by \SynSetExpan. 
We can see that \SynSetExpan is able to detect different types of entity synsets across various semantic classes. 
Furthermore, we highlight those entities discovered only after model fine-tuning, and we can see clearly that with fine-tuning, our \SynSetExpan framework can detect more accurate synsets. 

%!TEX root = main.tex
% UTF-8 encoding
\section{Related Work}\label{sec:related_work}

\vspace{-0.2cm}
\noindent \textbf{Entity Set Expansion.}
Entity set expansion can benefit many downstream applications such as question answering~\cite{Wang2008IterativeSE}, literature search~\cite{Shen2018EntitySS}, and online education~\cite{Yu2019CourseCE}. 
Traditional entity set expansion systems such as GoogleSet \cite{tong2008system} and SEAL \cite{Wang2007LanguageIndependentSE} 
require seed-oriented online data extraction, which can be time-consuming and costly. 
Thus, more recent studies~\cite{Shen2017SetExpanCS, Mamou2018TermSE, Yu2019CorpusbasedSE, Huang2020GuidingCS, Zhang2020ESELM} are proposed to expand the seed set by offline processing a given corpus. 
These corpus-based methods include two general approaches: 
(1) \emph{one-time entity ranking}~\cite{Pantel2009WebScaleDS, He2011SEISASE, Mamou2018TermSE,Kushilevitz2020ATM} which calculates all candidate entities' distributional similarities with seed entities and makes a one-time ranking without back and forth refinement, and 
(2) \emph{iterative bootstrapping}~\cite{Rong2016egoset, Shen2017SetExpanCS, Huang2020GuidingCS, Zhang2020ESELM} which starts from seed entities to extract quality textual patterns; applies the extracted patterns to obtain more quality entities, and iterates this process until sufficient entities are discovered. 
In this work, in addition to just adding entities into the set, we go beyond one step and aim to organize those expanded entities into synonym sets. 
Furthermore, we show those detected synonym sets can in turn help to improve set expansion results. 

\vspace{0.1em}
\noindent \textbf{Synonym Discovery.}
Early efforts on synonym discovery focus on finding entity synonyms from structured or semi-structured data such as query logs~\cite{Ren2015SynonymDF}, web tables~\cite{He2016AutomaticDO}, and synonymy dictionaries~\cite{Ustalov2017WatsetAI, Ustalov2017FightingWT}.
In comparison, this work aims to develop a method to extract synonym sets directly from raw text corpus. 
Given a corpus and a term list, one can leverage surface string~\cite{Wang2019SurfConSD}, co-occurrence statistics~\cite{Baroni2004UsingCS}, textual pattern~\cite{Yahya2014ReNounFE}, distributional similarity~\cite{Wang2015SolvingVC}, or their combinations~\cite{Qu2017AutomaticSD, Fei2019HierarchicalMW} to extract synonyms.
These methods mostly find synonymous \emph{term pairs} or \emph{a rank list} of query entity's synonym, instead of entity synonym sets.
Some studies propose to further cut-off the rank list into a set output~\cite{Ren2015SynonymDF} or to build a synonym graph and then apply graph clustering techniques to derive synonym sets~\cite{Oliveira2014ECOAO,Ustalov2017WatsetAI}. 
However, they all operate directly on the entire input vocabulary which can be too extensive and noisy. 
Comparing to them, our approach can leverage the semantic class information detected from set expansion to enhance the synonym set discovery process.

%!TEX root = main.tex
% UTF-8 encoding
\section{Conclusions}\label{sec:conclusion}
\vspace{-0.1cm}
This paper shows entity set expansion and synonym discovery are two tightly coupled tasks and can mutually enhance each other.
We present \SynSetExpan, a novel framework jointly conducting two tasks, and \Dataset dataset, the first large-scale synonym-enhanced set expansion dataset.
Extensive experiments on \Dataset and several other benchmark datasets demonstrate the effectiveness of \SynSetExpan on both tasks.
In the future, we plan to study how we can apply \SynSetExpan at the entity mention level for conducting contextualized synonym discovery and set expansion. 
%!TEX root = main.tex
% UTF-8 encoding

\section*{Acknowledgements}
Research was sponsored in part by US DARPA KAIROS Program No. FA8750-19-2-1004 and SocialSim Program No. W911NF-17-C0099, NSF IIS 16-18481, IIS 17-04532, and IIS 17-41317, and DTRA HDTRA11810026. 
Any opinions, findings or recommendations expressed herein are those of the authors and should not be interpreted as necessarily representing the views, either expressed or implied, of DARPA or the U.S. Government. 
We thank anonymous reviewers for valuable and insightful feedback.

\bibliographystyle{acl_natbib}
\bibliography{cited}

%!TEX root = main.tex
% UTF-8 encoding

\clearpage

\appendix

%\section{Supplementary Materials}\label{appendix}

\section{Entity Pair Features}

\begin{table}[!h]
\centering
\scalebox{0.60}{
\begin{tabular}{c|c}
\toprule
\textbf{Feature Description} & \textbf{Example} \\
\midrule
IsPrefix & (Florida, FL) $\rightarrow$ 1 \\
\hline
IsInitial & (\underline{N}orth \underline{C}arolina, NC) $\rightarrow$ 1 \\
\hline
Edit distance & (North Carolina, Texas) $\rightarrow$ 13 \\
\hline
Jaro-Winkler similarity & (Arizona, Texas) $\rightarrow$ 0.4476 \\
\hline
Characters in common & (Lone Star State, Texas) $\rightarrow$ 2 \\
\hline
Tokens in common & (North \underline{Carolina}, South \underline{Carolina}) $\rightarrow$ 1 \\
\hline
Difference in \#tokens & (Land of Lincoln, Illinois) $\rightarrow$ |3-1| = 2 \\
\hline
Initial edit distance & (\underline{N}orth \underline{C}arolina, \underline{S}tate \underline{o}f \underline{N}orth \underline{C}arolina) $\rightarrow$ 2 \\
\hline
Longest token edit distance & (North \underline{Dakota}, North \underline{Carolina}) $\rightarrow$ 5 \\
\hline
Cosine similarity of embedding & (Texas, Lone Star State) $\rightarrow$  0.9 \\
\hline
Transformed cosine similarities & (Texas, Lone Star State) $\rightarrow  [\frac{1}{0.9}, \sqrt{0.9}, (0.9)^{2}] $ \\
\hline
Multiplication of two entities'  & (Illinois, Land of Lincoln) $\rightarrow$  \\
PCA-reduced embedding & [0.006, 0.072, -0.008, 0.074, $\cdots$, -0.004] \\
\bottomrule
\end{tabular}
}
\vspace*{-0.2cm}
\caption{\label{tbl:entity_pair_features} All entity pair features used in our synonym discovery model.}
\vspace*{-0.3cm}
\end{table}

\section{SynSetExpan Framework Complexity}
From the Algorithm 1 in the main text, we can see our \SynSetExpan framework costs $O(T\times(1+K)\times|S|+|\V|)$ for each iteration, where $T$ is the ensemble times (usually 50), $K$ is the negative sampling size (usually 10-20), $S$ is the currently expanded set (usually of size $<100$), and $|\V|$ is the vocabulary size. 
Although such complexity looks expensive, we can significantly reduce the practical running time in two ways.
First, we can learn $T$ separate classifiers in set expansion model in parallel.
Second, we can aggregate all words in the vocabulary into one batch and apply synonym discovery model for inference in one run. 
We report the practical running time for each component in the below experiments.

\section{Existing ESE Datasets}\label{subsec:dataset_comparision_details}
An ideal set expansion benchmark dataset should contain four parts: a corpus, a vocabulary, a set of complete semantic classes, and a collection of seed queries for each semantic class. 
One of the earliest corpus-based set expansion work~\cite{Pantel2009WebScaleDS} uses \mquote{List of} pages in Wikipedia to construct 50 semantic classes and applies random sampling to construct 30 queries for each class. 
Although those classes and queries are still available today, we have no access to its underlying corpus and vocabulary and thus cannot easily reproduce their results. 
Similarly, SEISA~\cite{He2011SEISASE} and EgoSet~\cite{Rong2016egoset} also release their constructed semantic classes and seed queries but hold the corpus and vocabulary. 
On the other side, SetExpander~\cite{Mamou2018TermSE} and CaSE~\cite{Yu2019EfficientCS} clearly describe their corpus and vocabulary but do not release their classes/queries. 
To the best of our knowledge, SetExpan~\cite{Shen2017SetExpanCS} is the only public dataset consisting of all four essential components.
However, it only contains 65 queries from 13 classes and has no synonym information. 
Below Table~\ref{tbl:dataset_comparison} compares our proposed \Dataset with all existing datasets and we can see that our new dataset contains all four key parts for a set expansion benchmark dataset, as well as additional synonym information. 

% Table: Dataset Comparison.
\begin{table}[!th]
    \centering
\scalebox{0.53}{
	\begin{tabular}{c|ccccc}
        \toprule
        \textbf{Dataset}    & Corpus &  Vocab & Classes & Queries & Synonyms \\
        \midrule
        \textbf{Pantel \emph{et al.}} \cite{Pantel2009WebScaleDS} & $\times$   & $\times$ & $\checkmark$ & $\checkmark$  & $\times$ \\
        \textbf{SEISA} \cite{He2011SEISASE}        & $\times$   & $\times$ & $\checkmark$ & $\checkmark$  & $\times$ \\
        \textbf{EgoSet} \cite{Rong2016egoset} & $\times$   & $\times$ & $\checkmark$ & $\checkmark$  & $\times$ \\
        \textbf{SetExpander} \cite{Mamou2018TermSE} & $\checkmark$   & $\checkmark$ & $\times$ & $\times$  & $\times$ \\
        \textbf{CaSE} \cite{Yu2019EfficientCS} & $\checkmark$   & $\checkmark$ & $\times$ & $\times$  & $\times$ \\
        \textbf{SetExpan} \cite{Shen2017SetExpanCS} & $\checkmark$   & $\checkmark$ & $\checkmark$ & $\checkmark$  & $\times$ \\
        \midrule
        \DatasetBf & $\checkmark$   & $\checkmark$ & $\checkmark$ & $\checkmark$  & $\checkmark$ \\
        \bottomrule
        \end{tabular}
}
	\vspace*{-0.2cm}
	\caption{Comparison of ESE datasets.}\label{tbl:dataset_comparison}
	\vspace*{-0.3cm}
\end{table}

\section{\Dataset Dataset Construction Details}\label{subsec:dataset_construction_details}

We construct our dataset in four stages: (1) Corpus and vocabulary selection, (2) Semantic class selection, (3) Query generation and class enrichment, and (4) Synonym set curation.

\smallskip
\noindent \textbf{Corpus and vocabulary selection.} 
An ideal corpus for set expansion task should contain a diverse set of semantic classes and enough context information for methods to discover those sets.
Based on these two criteria, we select Wikipedia 20171201 dump as our evaluation corpus.
This corpus is also used in previous studies \cite{Mamou2018TermSE, Mamou2018TermSE2} and contains 1.9 billion tokens of raw size 14GB. 
Next, we extract all noun phrases with frequency above 10 and filter out those noun phrases that start with either a stopword (\eg, \mquote{a/an} and \mquote{the}) or a non-word character (\eg, \mquote{(}, and \mquote{-}). 
The remaining 1.47 million noun phrases consist of our vocabulary. 

\smallskip
\noindent \textbf{Semantic class selection.} 
To select a diverse set of semantic classes, we first use simple string matching to align our corpus and vocabulary with two benchmark datasets designed for tasks closely related to Set Expansion: (1) \emph{DBpedia-Entity v2} \cite{Hasibi2017DBpediaEntityVA} for Entity Search (particularly entity list search), and (2) \emph{WikiTable} \cite{Bhagavatula2015TabELEL, Zhang2018OntheflyTG} for Entity Linking in Wikipedia Table. 
Then, we retain all semantic classes with at least 10 entities and obtain totally 60 classes covering 6 different types (\eg, Person, Location, Organization, etc). 
Table~\ref{tbl:dataset_example} shows some examples.
Such generated classes have high precision but low recall in the sense that some correct entities are not included. 
In the following stage, we enlarge each semantic class and increase its coverage using crowdsourcing. 

\smallskip
\noindent \textbf{Query generation and class enrichment.} 
For each semantic class, we generate 5 queries for each of four query sizes: 2, 3, 4, 5, which results in 20 queries per class and 1200 queries in total.  
Furthermore, we want those queries to cover both popular and long-tail entities. 
To achieve this goal, we first sort all entities based on their frequencies within each class.
Then, we generate each subgroup of 5 queries (of the same size $M \in \{2, 3, 4, 5\}$) as follows: we select 1 query consisting of the $M$ most frequent entities, 2 queries of entities in frequency quantile top-10\%, and 2 queries of entities in frequency quantile [top-10\%, top-30\%]. 

After generating queries, we run all baseline methods to retrieval their top 100 results and aggregate all results to a set of 17,400 $\pair{\text{class}, \text{entity}}$ pairs. 
Next, we employ crowdworkers to check all those pairs on Amazon Mechanical Turk.
Crowdworkers are required to have a 95\% HIT acceptance rate, a minimum of 1000 HITs, and be located in the United States or Canada. 
Workers are asked to view one semantic class and six candidate entities, and to select all entities that belong to the given class. 
On average, workers spend 40 seconds on each task and are paid \$0.1, which is equivalent to a \$9 hourly payment. 
All $\pair{\text{class}, \text{entity}}$ pairs are labeled by three workers independently and the inter-annotator agreement is 0.8204, measured by Fleiss's Kappa ($k$).
Finally, we enrich each semantic class $C_j$ by adding the entity $e_i$ whose corresponding pair $\pair{C_j, e_i}$ is labeled ``True'' by at least two workers. 

\smallskip
\noindent \textbf{Synonym set curation.} 
To construct synonym sets in each semantic class, we first run all baseline methods to generate a candidate pool of possible synonymous pairs. 
Then, we enlarge this pool to include all term pairs that form an inflection\footnote{\small We check word inflection using: \url{https://github.com/jazzband/inflect}.}. 
After that, we automatically treat those terms that can be mapped to the same entity in WikiData\footnote{\small \url{https://www.wikidata.org/wiki/Wikidata:Main_Page}} as positive pairs and manually label the remaining 7,625 pairs.
The inter-annotator agreement is 0.8431.
Note here we do not use Amazon MTurk because labeling synonym pairs are much simpler than labeling entity class membership and also has less ambiguity. 
Here, we avoid using YAGO KB in order to prevent possible data leakage problem.
Next, we construct a synonym graph where each node is a term and each edge connects two synonymous terms. 
Finally, we extract all connected components in this synonym graph and treat them as synonym sets. 

%% Table: Dataset Samples
\begin{table*}[!thpb]
\centering
\scalebox{0.68}{
%\begin{tabular}{|l|l|l|l|c|}
\begin{tabular}{l|l|l|l}
\toprule
\textbf{Class ID} & \textbf{Class Name} & \textbf{Class Type (Class Description)} & \textbf{Entities with Synsets}  \\
\midrule
\multirow{2}{*}{WikiTable-21} & \multirow{2}{*}{U.S. states} & \multirow{2}{*}{LOC (Locations)}  & [\{\mquote{Texas}, \mquote{TX}, \mquote{Lone Star State}\}, \{\mquote{Arizona}, \mquote{AZ}\},  \\
 & &  & ~~~~\{\mquote{California}, \mquote{CA}, \mquote{Golden State}\}, ......] \\
\hline
\multirow{2}{*}{SemSearch-LS-3} & Astronauts who landed & \multirow{2}{*}{PERSON (People)}  & [\{\mquote{Eugene Andrew Cernan}, \mquote{Gene Cernan}\}, \{\mquote{Pete Conrad}\}, \\
			   & on the Moon &  & ~~~~\{\mquote{Neil A. Armstrong}, \mquote{Neil Armstrong}\}, ......] \\
\hline
Enriched-1 & Apple Products & PRODUCT (Objects, vehicles, ...)  & [\{\mquote{MacBook Pro}, \mquote{MBP}\}, \{ \mquote{iTouch}, \mquote{iPod Touch}\}, ......] \\
\hline
\multirow{2}{*}{Enriched-3} & \multirow{2}{*}{Volcanoes in USA} & \multirow{2}{*}{LOC (Non-GPE locations)}  & [\{\mquote{Yellowstone}\}, \{\mquote{Mount Rainier}, \mquote{Tahoma}, \mquote{Tacoma}\}, \\
 & &  & ~~~~\{\mquote{Mount Hood}, \mquote{Mt. Hood}, \mquote{Wy'east}\}, ......] \\
\hline
\multirow{2}{*}{WikiTable-27} & \multirow{2}{*}{Airports in British Isles} & \multirow{2}{*}{FAC (Facilities)}  &  [\{\mquote{Ringway Airport}, \mquote{Manchester Airport} \}, \\
 &  &  & ~~~~\{\mquote{RAF Exeter}, \mquote{Exeter International Airport}\}, ......] \\
\hline
\multirow{2}{*}{Enriched-4} & \multirow{2}{*}{NBA Teams} & \multirow{2}{*}{ORG (Organizations)}  & [\{\mquote{Washington Bullets}, \mquote{Washington Wizards} \}, \\
 &  &   & ~~~~\{\mquote{Los Angeles Lakers}, \mquote{L.A. Lakers}, \mquote{Lakers}\}, ......] \\
\hline
\multirow{2}{*}{INEX-XER-147} & Chemical elements that  & \multirow{2}{*}{MISC (Miscellaneous classes)}  & [\{\mquote{Gadolinium}\}, \{\mquote{Seaborgium}, \mquote{Element 106}\},  \\
			& are named after people &  & ~~~~\{\mquote{Einsteinium}, \mquote{Es99}\}, ......] \\
\bottomrule
\end{tabular}}
\vspace*{-0.2cm}
\caption{\label{tbl:dataset_example} Example Semantic Classes in \DatasetBf Dataset.}
\vspace*{-0.3cm}
\end{table*}

\section{SE2 Dataset Analysis}

We analyze some properties of \Dataset dataset from the following aspects: (1) semantic class size, (2) set expansion difficulty of each class, and (3) synonym discovery difficulty of each class. 

\smallskip
\noindent \textbf{Semantic class size.} 
The 60 semantic classes in our \Dataset dataset consist on average 145 entities (with a minimum of 16 and a maximum of 864) for a total of 8697 entities. 
After we grouping these entities into synonym sets, these 60 classes consist of on average 118 synsets (with a minimum of 14 and a maximum of 800) for totally 7090 synsets. 
The average synset size is 1.258 and the maximum size of one synset is 11. 

\smallskip
\noindent \textbf{Set expansion difficulty of each class.} 
We define the set expansion difficulty of each semantic class as follows:
\begin{equation}\label{eq:set_expansion_difficulty}
\small
\text{Set-Expansion-Difficulty}(C) = \frac{1}{|C|}\sum_{e \in C} \frac{|C - \text{Top}k(e)|}{|C|},
\end{equation}
where Top$k(e)$ represents the set of $k$ most similar entities to entity $e$ in the vocabulary. 
We set $k$ to be 10,000 in this study. 
Intuitively, this metric calculates the average portion of entities in class $C$ that cannot be easily found by another entity in the same class. 
As shown in Table~\ref{tbl:data_analysis}, the most difficult classes are those LOC classes\footnote{\small We exclude MISC type because by its definition classes of this type will be very random.} and the easiest ones are FAC classes. 

\smallskip
\noindent \textbf{Synonym discovery difficulty of each class.} 
We continue to measure the difficulty of finding synonym pairs in each class.
Specifically, we calculate two metrics: (1) \emph{Lexical difficulty} defined as the average Jaro-Winkler distance\footnote{\small We use Jaro-Winkler distance instead of other edit distances because it is symmetric, normalized to range from 0 to 1, and is widely used in previous synonym literature.} between the surface names of two synonyms, and (2) \emph{Semantic difficulty} defined as the average cosine distance between two synonymous entities' embeddings. 
Table~\ref{tbl:data_analysis} lists the results. 
We find PRODUCT classes have the largest lexical difficulty and LOC classes have the largest semantic difficulty. 

% Table: Dataset Analysis.
\begin{table}[!t]
\centering
\scalebox{0.75}{
\begin{tabular}{cccc}
\toprule
\textbf{Class Type} & \textbf{ESE} & \textbf{ESD (Lexical)} & \textbf{ESE (Semantic)} \\
\midrule
Location & 0.3789 & 0.2132 & 0.6599 \\
Person & 0.2322 & 0.2874 & 0.5526 \\
Product & 0.0848 & 0.3922 & 0.4811 \\
Facility &0.0744 & 0.2345 & 0.4466 \\
Organization & 0.1555 & 0.2566 & 0.4935 \\ 
Misc & 0.4282 & 0.2743 & 0.5715 \\  
\bottomrule
\end{tabular}}
\vspace*{-0.1cm}
\caption{\label{tbl:data_analysis2}Difficulty of each semantic class for entity set expansion (ESE) and entity synonym discovery (ESD).}
\vspace*{-0.3cm}
\end{table}

\section{Entity Set Expansion Experiments}\label{sec:set_expan_details}

\subsection{Implementation Details and Hyper-parameter Choices}

For Wiki and APR datasets, we directly report each baseline method's performance obtained in the CGExpan paper~\cite{Zhang2020ESELM}.
For our proposed SE2 dataset, we tune each method's hyper-parameters on 6 semantic classes (one for each class type) and use tuned parameters for all the other classes. 
The implementation details and specific hyper-parameter choices are discussed below:
\begin{enumerate}[leftmargin=*]
\item \textbf{EgoSet}: There is no open-source code for EgoSet and thus we implement it on our own. We use each entity's 250 most relevant skip-grams to calculate entity-entity similarity.
\item \textbf{SetExpan}\footnote{\small \url{https://github.com/mickeystroller/SetExpan}}: We run SetExpan for 10 iterations and add 10 entities into the set in each iteration. We set ensemble time to be 90 and use the default values for all other hyper-parameters.
\item \textbf{SetExpander}\footnote{\small \url{http://nlp_architect.nervanasys.com/term_set_expansion.html}}: We directly download the pre-trained vectors (as they are trained on the same corpus as ours) and filter out those words that do not exist in our vocabulary.
\item \textbf{MCTS}\footnote{\small \url{https://github.com/lingyongyan/mcts-bootstrapping}}: In each iteration, we perform 1000 MCTS simulations and select 10 patterns to add 10 entities. 
\item \textbf{CaSE}\footnote{\small \url{https://github.com/PxYu/entity-expansion}}: We use its CaSE-BERT version where a BERT-base-uncased model is used to calculate entity representations.
\item \textbf{CGExpan}\footnote{\small \url{https://github.com/yzhan238/CGExpan}}: We use BERT-base-uncased as its underlying Language Model for generating class names. We run CGExpan for 5 iterations and each iteration finds 5 candidate classes and adds 10 most confident entities into the currently expanded set. 
\item \SynSetExpanBf: We set the ensemble times $T=50$, the negative sampling ratio (in set expansion model) $K=10$, the maximum iteration number \emph{max$\underline{\hspace{0.05in}}$iter}$=6$, the number of fine-tuning trees $H=10$, and the negative sampling ratio (in synonym discovery model) $N=10$. For other (less important) hyper-parameters, we directly discuss their values in the paper and \SynSetExpan is robust to those hyper-parameters.
\end{enumerate}

\subsection{Hyper-parameter Sensitivity Analysis}\label{subsec:sensitvy}
\begin{figure}[!t]
\subfigure[Ensemble Times $T$]{
\includegraphics[width = 0.22\textwidth]{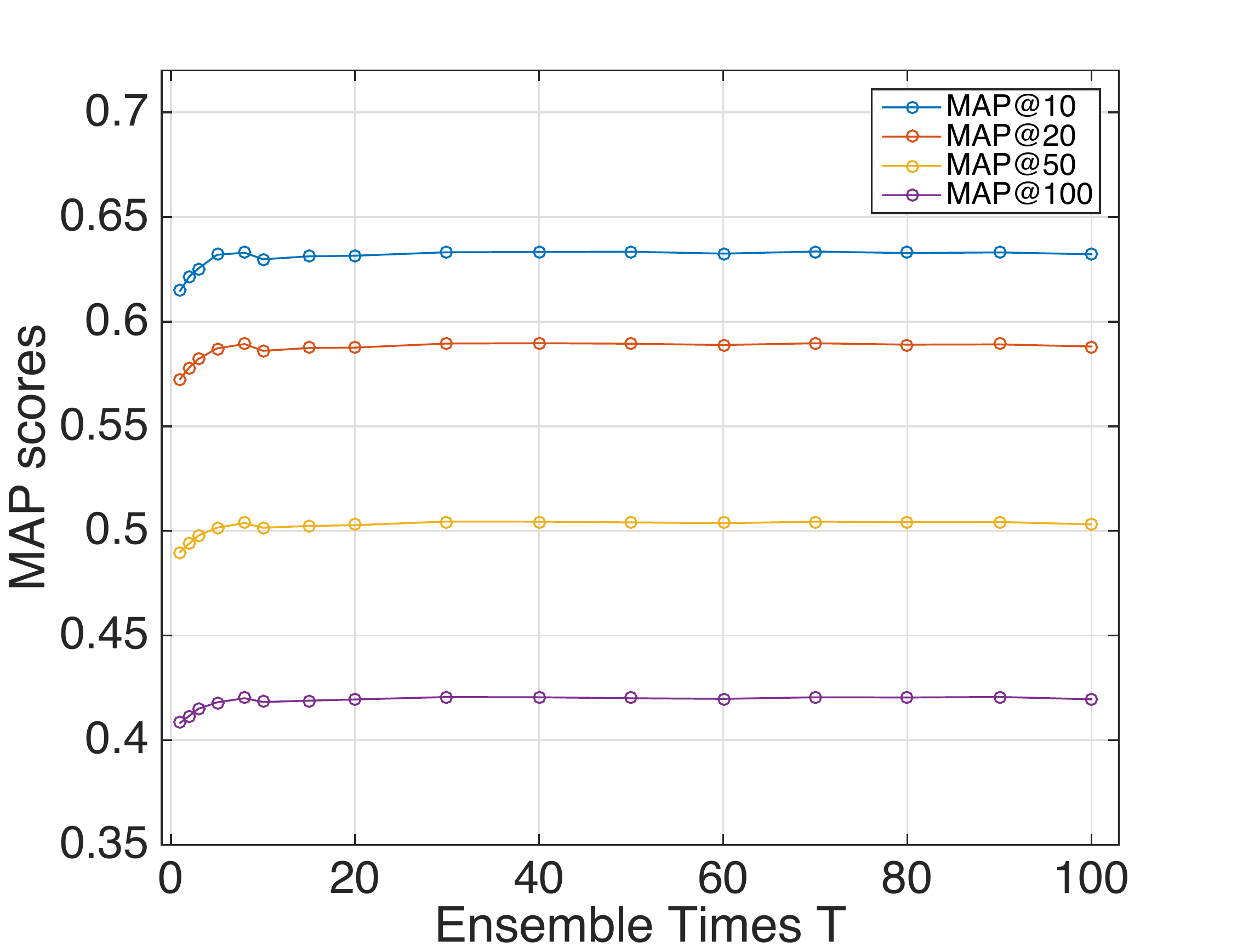}
}
\subfigure[Negative Ratio $K$]{
\includegraphics[width = 0.22\textwidth]{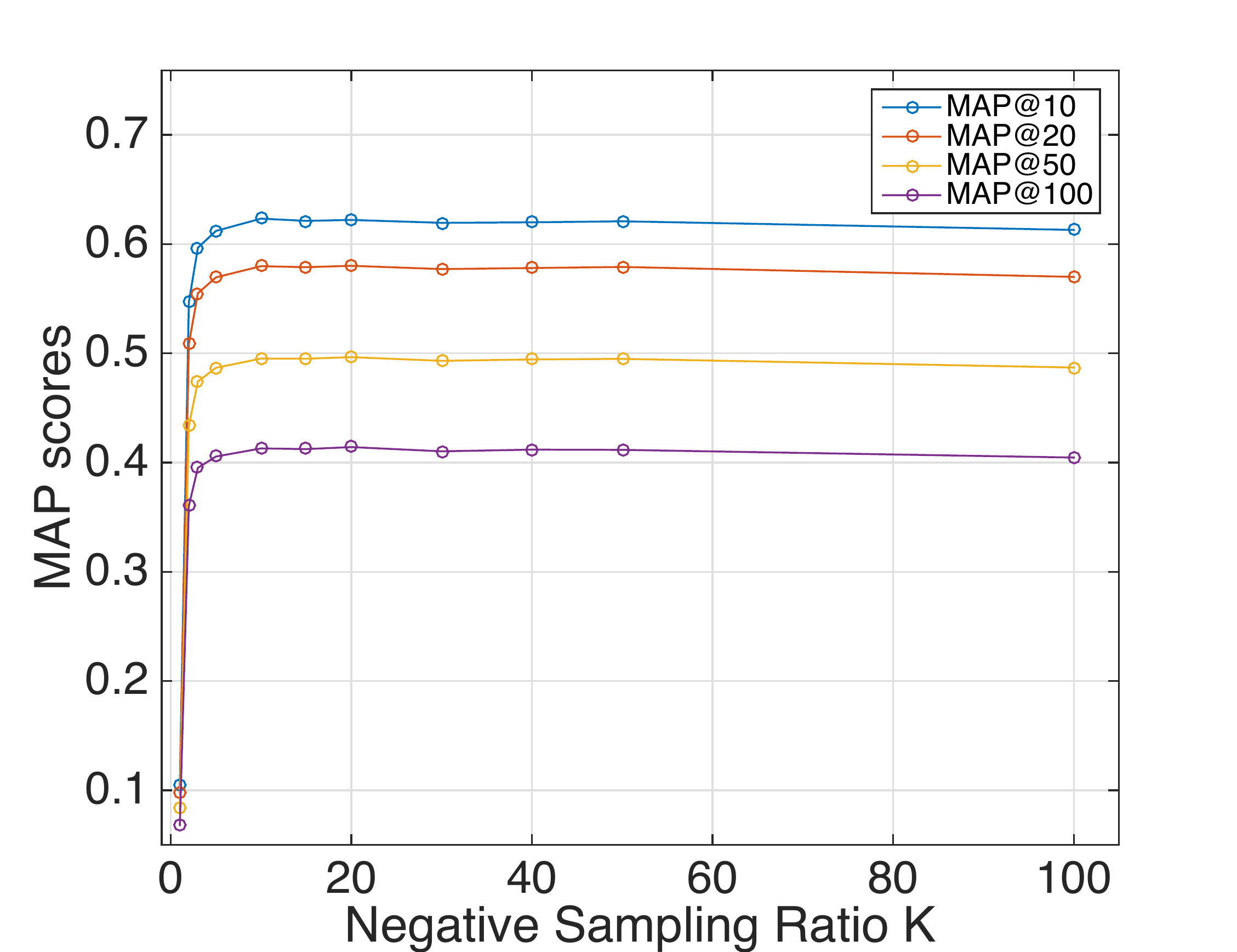}
}
\vspace{-0.3cm}
\caption{Sensitivity analysis of hyper-parameters $T$ and $K$ in \SynSetExpan for the set expansion task.}
\label{fig:setexpan_parameter_sense}
\vspace{-0.3cm}
\end{figure}

Within our \SynSetExpan framework, there are two important hyper-parameters in the set expansion model: the ensemble times $T$ and negative sampling ratio $K$.
Figure~\ref{fig:setexpan_parameter_sense} shows the hyper-parameter sensitivity analysis.
We see that the model performance first increases when the ensemble times $T$ increases from 1 to 10 and then becomes stable when $T$ further increases. 
A similar trend is also witnessed on the negative sampling ratio $K$.
Overall, we can say that SynSetMine is insensitive to these two hyper-parameters as long as their values are larger than 10. 

\subsection{Efficiency Analysis}\label{subsec:setexpan_efficiency}

We test the efficiency of our SynSetExpan framework (with $T$ = 50 and $K$ = 10) on a single server with 20 CPU threads. 
For each query, the first iteration of SynSetExpan on average takes 7.5 seconds, the first three iterations need 27 seconds, and the first six iterations consume 56 seconds. 
Later iterations take longer time because there are more entities in the already expanded set of that iteration.
In comparison, one iteration of EgoSet takes 86 seconds, six iterations of SetExpan need 188 seconds, and CGExpan takes 174 seconds for five iterations on a 1080Ti GPU.
This result shows the efficiency of SynSetExpan.

\section{Synonym Discovery Experiments}\label{sec:synset_discovery_details}

%% Figure: Hyper-parameter Sensitivity Analysis for Set Expansion Task

\subsection{PubMed Dataset Details}\label{sec:pubmed_details}

Besides using our \Dataset dataset, we also evaluate \SynSetExpan for synonym discovery task on the public PubMed dataset which consists of a corpus of 1.5 million paper abstracts in biomedical domain, a vocabulary of 357,991 terms, and a collection of 203,648 synonym pairs (10,486 positive pairs and 193,162 negative pairs). 
All terms involved in synonym pairs are linked to one entity in UMLS knowledge base\footnote{\scriptsize \url{https://uts.nlm.nih.gov/home.html}} and we group these terms into 10 semantic classes based on their linked entities' types.

\subsection{Implementation Details and Hyper-parameter Choices}\label{sec:syn_discovery_details}

All compared synonym discovery methods are tested using the same distant supervision data (c.f. Section 3 in the main text) and hyper-parameter values are obtained using 5-fold cross validation.
We discuss the implementation details and hyper-parameter choices of each compared synonym discovery methods below:
\begin{enumerate}[leftmargin=*]
\item \textbf{SVM}\footnote{\small \url{https://scikit-learn.org/stable/modules/generated/sklearn.svm.SVC.html\#sklearn.svm.SVC}}: We use the  RBF kernel and set regularization parameter $\lambda$ to be 0.3.
\item \textbf{XGBoost}\footnote{\small \url{https://github.com/dmlc/xgboost}}: We set the maximum tree depth to be 5, $\gamma=0.1$, $\eta=0.1$, subsample ratio to be 0.5, and use the default values for all other hyper-parameters.
\item \textbf{SynSetMine}\footnote{\small \url{https://github.com/mickeystroller/SynSetMine-pytorch}}: We use two hidden layers (of dimension 250, 500) for its internal set encoder. We learn the model using the ``mix sampling'' strategy.
\item \textbf{DPE}\footnote{\small \url{https://github.com/mnqu/DPE}}: We set the embedding dimension as 300, $\lambda = 0.3$, and use the default values for all other hyper-parameters.
\item \SynSetExpanBf: We use the same hyper-parameter values as XGBoost to obtain the class-agonistic synonym discovery model. During the fine-tuning stage, we fit $10$ additional trees in each iteration. For other (less important) hyper-parameters, we directly discuss their values in the paper and \SynSetExpan is robust to those hyper-parameters.
\end{enumerate}

%% Figure: Hyper-parameter Sensitivity Analysis for Synonym Discovery Task
\begin{figure}[!t]
\subfigure[\Dataset Dataset-$H$]{
\includegraphics[width = 0.22\textwidth]{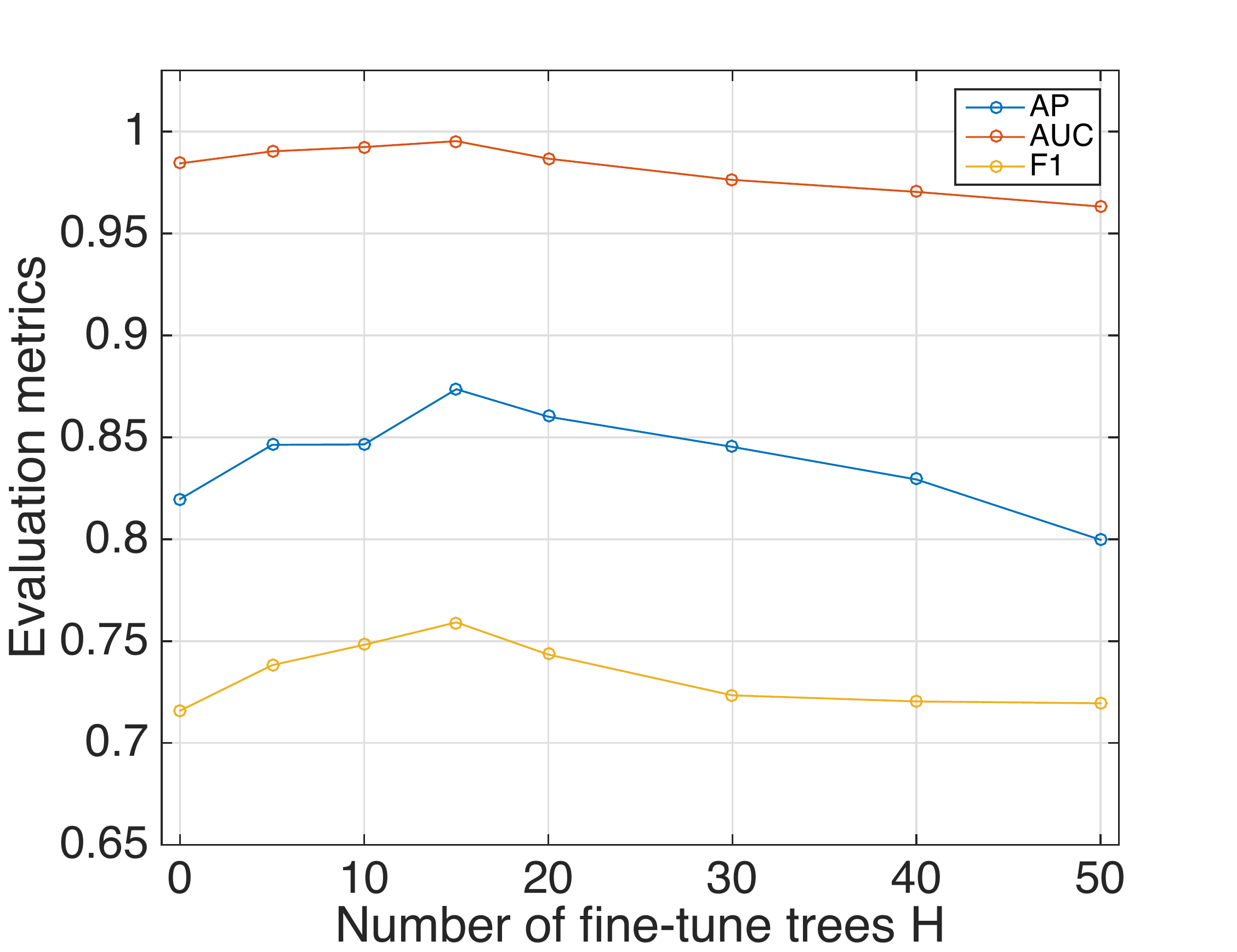}
}
\subfigure[\Dataset Dataset-$N$]{
\includegraphics[width = 0.22\textwidth]{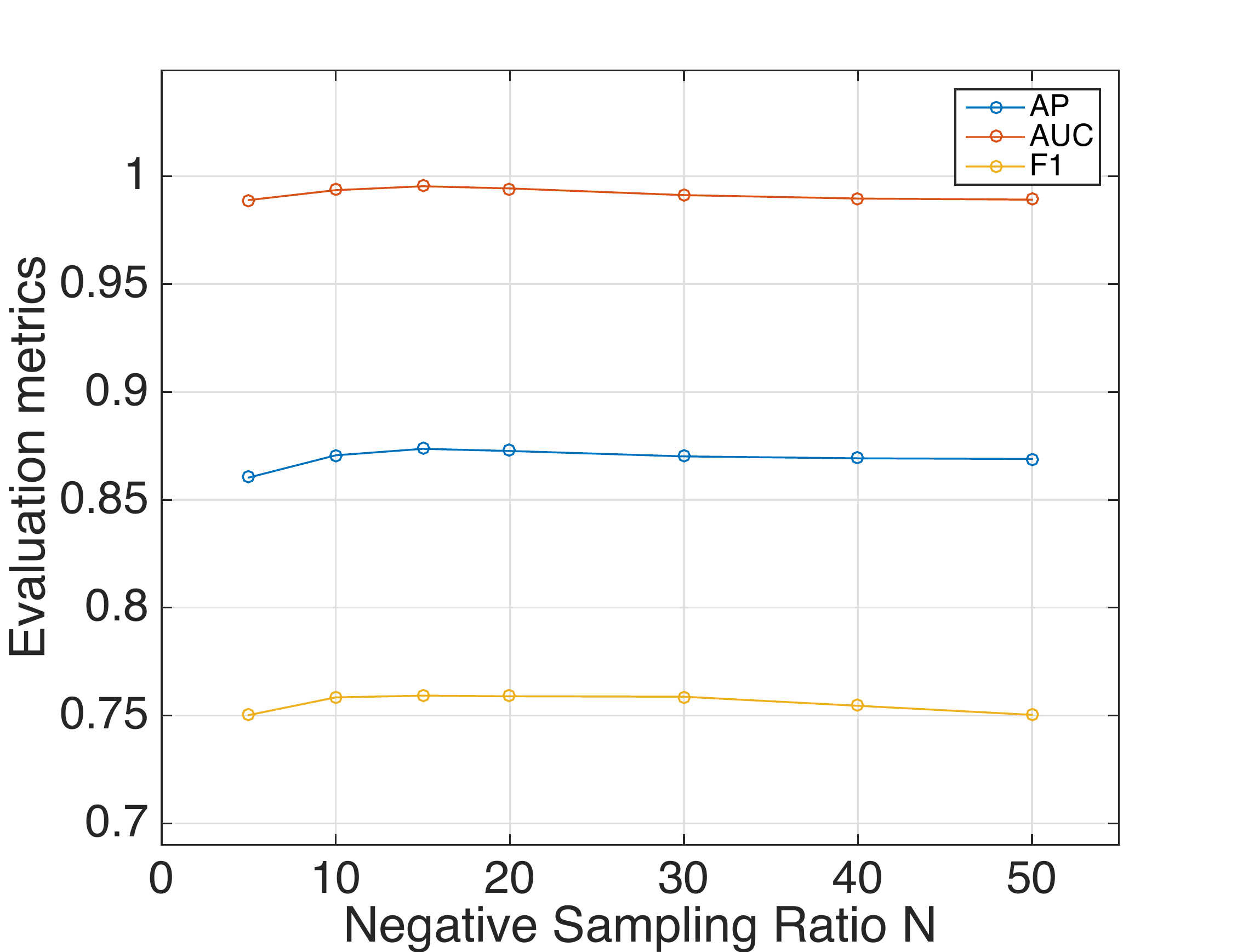}
}
\subfigure[PubMed Dataset-$H$]{
\includegraphics[width = 0.22\textwidth]{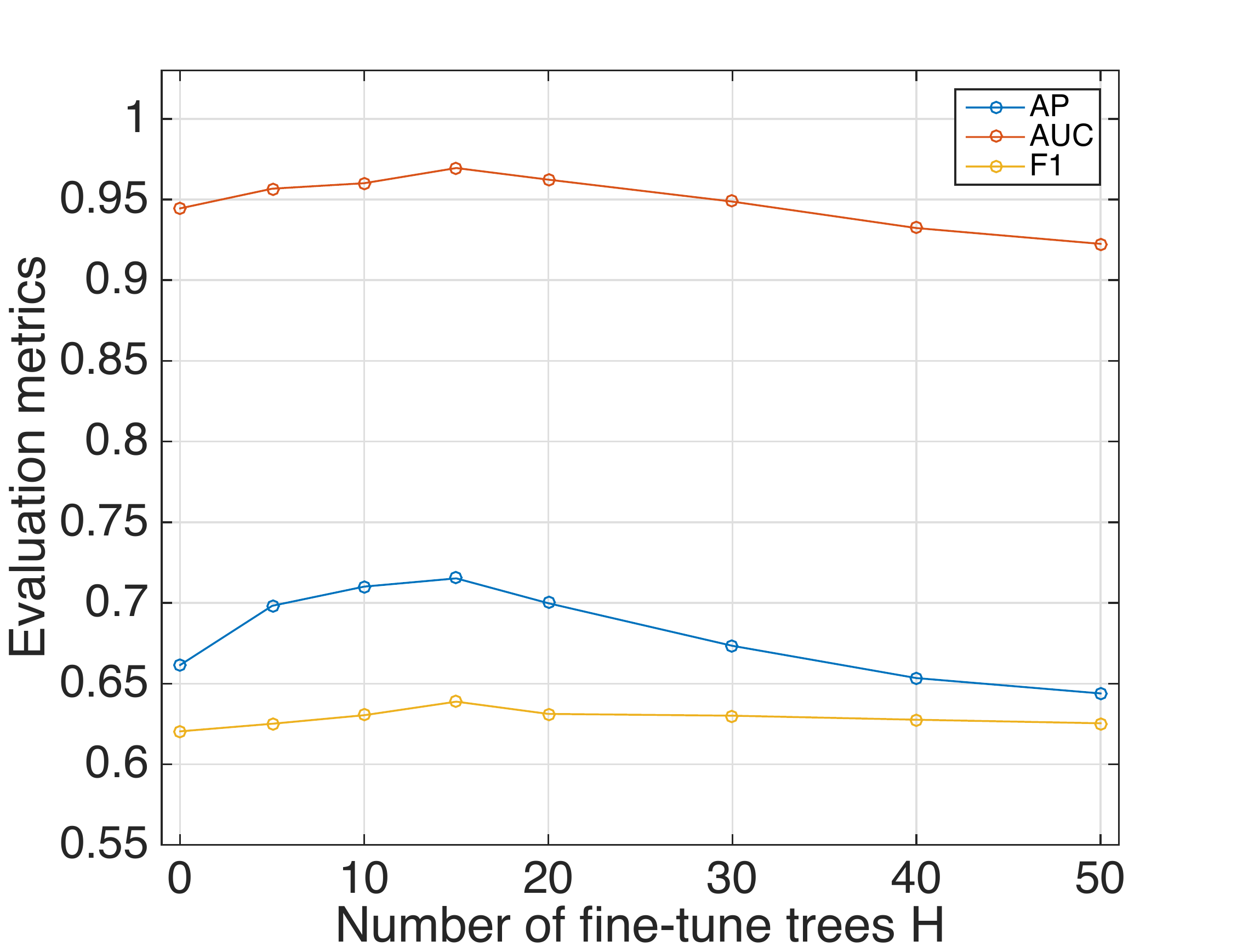}
}
\subfigure[PubMed Dataset-$N$]{
\includegraphics[width = 0.22\textwidth]{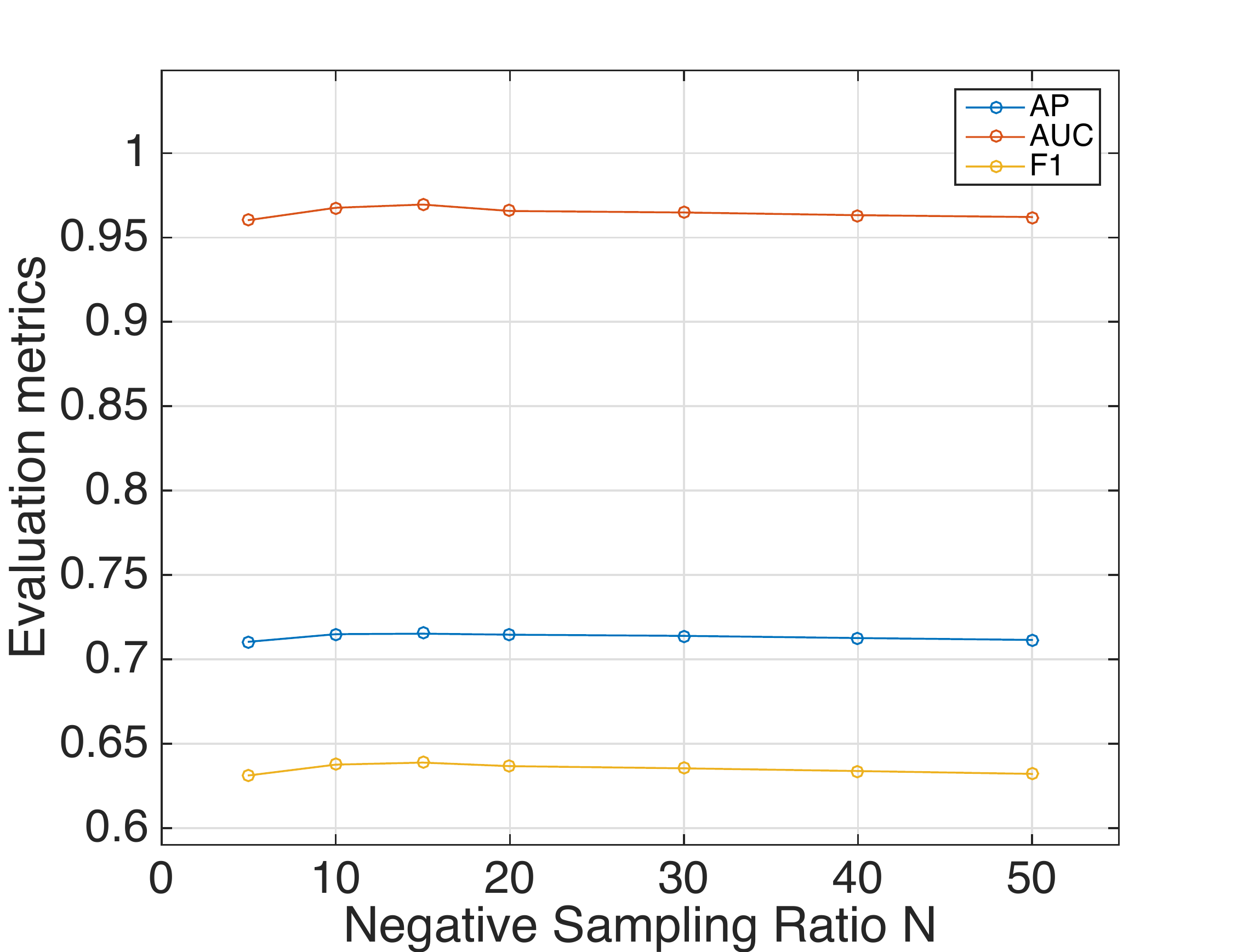}
}
\vspace{-0.2cm}
\caption{Sensitivity analysis of hyper-parameters $H$ and $N$ in \SynSetExpan framework for the synonym discovery task.}
\label{fig:synonym_parameter_sense}
\vspace{-0.3cm}
\end{figure}

\subsection{Hyper-parameter Sensitivity Analysis}

We study how sensitive \SynSetExpan is to the choices of two fine-tuning hyper-parameters in its synonym discovery module: (1) the number of additional fitted trees $H$, and (2) the negative sampling ratio $N$ in constructing pseudo-labeled dataset for fine-tuning. 
Results are shown in Figure~\ref{fig:synonym_parameter_sense}. 
First, we find that our model is insensitive to the negative sampling ratio $N$ in terms of all three metrics. 
Second, we notice that the model performance first increases as $H$ increases until it reaches about 15 and then starts to decrease when we further increase $H$.
Although \SynSetExpan is somewhat sensitive to the hyper-parameter $H$, we find that a wide range of $H$ choices are better than $H=0$ which essentially disables the model fine-tuning.

\subsection{Efficiency Analysis}
By linking \Dataset~Dataset with YAGO KB, we can obtain 260 thousand synonym pairs based on which training a class-agnostic synonym discovery model takes 15 minutes. 
Then, each iteration of \SynSetExpan generates on average 5000 pseudo-labeled synonym pairs and fitting 10 additional trees needs about 0.75 seconds. 
After training, our synonym discovery model can predict 4000 term pairs per second.

\end{document}